\documentclass[11pt,cls,onecolumn]{IEEEtran}
\usepackage[latin1]{inputenc}
\usepackage[T1]{fontenc}
\usepackage{cite}

\ifCLASSINFOpdf
   \usepackage[pdftex]{graphicx}
\else
\fi
\usepackage[cmex10]{amsmath}
\usepackage{amssymb}
\usepackage{algorithmic}
\usepackage{algorithm,bm}
\usepackage{enumerate}

\usepackage{color}
\usepackage{mdwmath}
\usepackage{mdwtab}
\usepackage[caption=false,font=footnotesize]{subfig}

\DeclareMathOperator{\tr}{tr}
\DeclareMathOperator{\rk}{rk}

\newtheorem{lemma}{Lemma}

\newcommand{\Vector}[1]{\mathbf{#1}}

\newcommand{\R}[1]{\mathbb{R}^{#1}}
\newcommand{\OP}{\mathbf{\Omega}}
\newcommand{\OPF}{\mathbf{\Omega}^F}
\newcommand{\OPo}{\mathbf{\Omega}}

\newcommand{\PSNR}{\textit{PSNR}}
\newcommand{\SSIM}{\textit{MSSIM}}

\makeatletter
\newlength \figwidth
\if@twocolumn
\else
\fi
\makeatother

\makeatletter
\newcommand{\thickhline}{%
    \noalign {\ifnum 0=`}\fi \hrule height 1pt
    \futurelet \reserved@a \@xhline
}
\newcolumntype{"}{@{\hskip\tabcolsep\vrule width 1pt\hskip\tabcolsep}}
\makeatother

\begin{document}
%
\title{Analysis Operator Learning and Its Application to Image Reconstruction}
%
%
%

\author{Simon~Hawe,
        Martin~Kleinsteuber,
        and~Klaus~Diepold
\thanks{ Copyright (c) 2013 IEEE. Personal use of this material is permitted. However, permission to use this material for any other purposes must be obtained from the IEEE by sending a request to pubs-permissions@ieee.org. The paper has been published in IEEE Transaction on Image Processing 2013.} 
\thanks{IEEE Xplore: http://ieeexplore.ieee.org/xpl/articleDetails.jsp?tp=\&arnumber=6459595\&contentType=Early+Access+Articles\&queryText\%3DSImon+Hawe}   
\thanks{DOI: 10.1109/TIP.2013.2246175 }       
\thanks{The authors are with the Department
of Electrical Engineering, Technische Universit\"at M\"unchen, Arcisstra\ss e 21, Munich
80290, Germany (e-mail: \{simon.hawe,kleinsteuber,kldi\}@tum.de, web: www.gol.ei.tum.de)}
\thanks{This work has been supported by the Cluster of Excellence \emph{CoTeSys} - Cognition for Technical Systems, funded by the German Research Foundation (DFG).}

}

%
%

\markboth{Technical Report}%
{Hawe \MakeLowercase{\textit{et al.}}: Analysis Operator Learning and Its Application to Image Reconstruction}
%

\maketitle

\begin{abstract}
http://
Exploiting a priori known structural information lies at the core of many image reconstruction methods that can be stated as inverse problems. The synthesis model, which assumes that images can be decomposed into a linear combination of very few atoms of some dictionary, is now a well established tool for the design of image reconstruction algorithms. An interesting alternative is the analysis model, where the signal is multiplied by an analysis operator and the outcome is assumed to be sparse. This approach has only recently gained increasing interest. The quality of reconstruction methods based on an analysis model severely depends on the right choice of the suitable operator.

In this work, we present an algorithm for learning an analysis operator from training images. Our method is based on $\ell_p$-norm minimization on the set of full rank matrices with normalized columns. We carefully introduce the employed conjugate gradient method on manifolds, and explain the underlying geometry of the constraints. Moreover, we compare our approach to state-of-the-art methods for image denoising, inpainting, and single image super-resolution. Our numerical results show competitive performance of our general approach in all presented applications compared to the specialized state-of-the-art techniques.
\end{abstract}


\ifCLASSOPTIONpeerreview
\begin{center} \bfseries EDICS Category: TEC-FOR (TEC-RST, TEC-ISR).\end{center}
\fi
%
\IEEEpeerreviewmaketitle

\section{Introduction}\label{sec:intro}
\subsection{Problem Description}
\IEEEPARstart{L}{inear} inverse problems are ubiquitous in the field of image processing. Prominent examples are image denoising \cite{noise:portilla:2003}, inpainting \cite{ip:bertalmio:2000}, super-resolution \cite{sri:freeman:2002}, or image reconstruction from few indirect measurements as in Compressive Sensing \cite{CS:candes:2006a}. Basically, in all these problems the goal is to reconstruct an unknown image $\mathbf{s} \in \R{n}$ as accurately as possible from a set of indirect and maybe corrupted measurements $\mathbf{y} \in \R{m}$ with $n \geq m$, see \cite{ip:kirsch:1991} for a detailed introduction to inverse problems. Formally, this measurement process can be written as
\begin{align}\label{eq:measurements}
\mathbf{y} = \mathcal{A} \mathbf{s} + \mathbf{e},
\end{align}
where the vector $\mathbf{e} \in \R{m}$  models sampling errors and noise, and $\mathcal{A} \in \R{m \times n}$ is the measurement matrix modeling the sampling process. In many cases, reconstructing $\mathbf{s}$ by simply inverting Equation \eqref{eq:measurements} is ill-posed because either the exact measurement process and hence $\mathcal{A}$ is unknown as in blind image deconvolution, or the number of observations is much smaller compared to the dimension of the signal, which is the case in Compressive Sensing or image inpainting. To overcome the ill-posedness and to stabilize the solution, prior knowledge or assumptions about the general statistics of images can be exploited.

\subsection{Synthesis Model and Dictionary Learning}\label{sec:syn_dict}
One assumption that has proven to be successful in image reconstruction, cf. \cite{sp:elad:2010}, is that natural images admit a sparse representation $\mathbf{x} \in \R{d}$ over some dictionary $\mathcal{D} \in \R{n \times d}$ with $d \geq n$. A vector $\mathbf{x}$ is called sparse when most of its coefficients are equal to zero or small in magnitude. When $\mathbf{s}$ admits a sparse representation over $\mathcal{D}$, it can be expressed as a linear combination of only very few columns of the dictionary $\{\mathbf{d}_i\}_{i=1}^d$, called \emph{atoms}, which reads as
\begin{align}\label{eq:dict_approx}
\mathbf{s} = \mathcal{D} \mathbf{x}.
\end{align}
For $d > n$, the dictionary is said to be overcomplete or redundant.

Now, using the knowledge that \eqref{eq:dict_approx} allows a sparse solution, an estimation of the original signal in \eqref{eq:measurements} can be obtained from the measurements $\mathbf{y}$ by first solving
\begin{align}\label{eq:std_min_syn}
\mathbf{x}^\star = \operatorname*{arg~min}_{\mathbf{x} \in \R{d}} \  g(\mathbf{x}) \operatorname{~subject~to~} \|\mathcal{A}\mathcal{D}\mathbf{x}-\mathbf{y}\|_2^2 \leq \epsilon,
\end{align}
and afterwards synthesizing the signal from the computed sparse coefficients via $\mathbf{s}^\star=\mathcal{D}\mathbf{x}^\star$. Therein, $g: \R{d} \to \R{}$ is a function that promotes or measures sparsity, and $\epsilon \in \R{+}$  is an estimated upper bound on the noise power $\|\Vector{e}\|_2^2$. Common choices for $g$ include the $\ell_p$-norm
\begin{align}\label{eq:pnorm}
\|\mathbf{v}\|_p^p:=\sum_i|v_i|^p,
\end{align}
with $0 < p \leq 1$ and differentiable approximations of \eqref{eq:pnorm}. As the signal is synthesized from the sparse coefficients, the reconstruction model \eqref{eq:std_min_syn} is called the \emph{synthesis} reconstruction model \cite{l1l2:elad:2007}.

To find the minimizer of Problem \eqref{eq:std_min_syn}, various algorithms based on convex or non-convex optimization, greedy pursuit methods, or Bayesian frameworks exist that may employ different choices of $g$. For a broad overview of such algorithms, we refer the interested reader to \cite{lp:tropp:2010}. What all these algorithms have in common, is that their performance regarding the reconstruction quality severely depends on an appropriately chosen dictionary $\mathcal{D}$. Ideally, one is seeking for a dictionary where $\mathbf{s}$ can be represented most accurately with a coefficient vector $\mathbf{x}$ that is as sparse as possible. Basically, dictionaries can be assigned to two major classes: \emph{analytic dictionaries} and \emph{learned dictionaries}.

Analytic dictionaries are built on mathematical models of a general type of signal, e.g. natural images, they should represent. Popular examples include Wavelets \cite{wav:mallat:1989}, Bandlets\cite{band:pennec:2005}, and Curvlets \cite{curv:starck:2002} among several others, or a concatenation of various such bases/dictionaries. They offer the advantages of low computational complexity and of being universally applicable to a wide set of signals. However, this universality comes at the cost of not giving the optimally sparse representation for more specific classes of signals, e.g. face images.

It is now well known that signals belonging to a specific class can be represented with fewer coefficients over a dictionary that has been learned using a representative training set, than over analytic dictionaries. This is desirable for various image reconstruction applications as it readily improves their performance and accuracy \cite{dl:elad:2006,dl:mairal:2010,dl:zhou:2012}.
Basically, the goal is to find a dictionary over which a training set admits a maximally sparse representation. In contrast to analytic dictionaries, which can be applied globally to an entire image, learned dictionaries are small dense matrices that have to be applied locally to small image patches. Hence, the training set consists of small patches extracted from some example images. This restriction to patches mainly arises from limited memory, and limited computational resources.

Roughly speaking, starting from some initial dictionary the learning algorithms iteratively update the atoms of the dictionary, such that the sparsity of the training set is increased. This procedure is often performed via block-coordinate relaxation, which alternates between finding the sparsest representation of the training set while fixing the atoms, and optimizing the atoms that most accurately reproduce the training set using the previously determined sparse representation. Three conceptually different approaches for learning a dictionary became well established, which are probabilistic ones like \cite{dl:delgado:2003}, 
clustering based ones such as K-SVD \cite{dl:aharon:2006}, and recent approaches which aim at learning dictionaries with specific matrix structures that allow fast computations like \cite{dl:rubinstein:2010}. For a comprehensive overview of dictionary learning techniques see \cite{dl:tosic:2011}.

\subsection{Analysis Model}
An alternative to the synthesis model \eqref{eq:std_min_syn} for reconstructing a signal, is to solve
\begin{align}\label{eq:std_min_an}
\mathbf{s}^\star = \operatorname*{arg~min}_{\mathbf{s} \in \R{n}} \ & g(\OP\mathbf{s}) \operatorname{~subject~to~} \|\mathcal{A}\mathbf{s}-\mathbf{y}\|_2^2 \leq \epsilon,
\end{align}
which is known as the \emph{analysis model} \cite{l1l2:elad:2007}. Therein, $\OP \in \R{k \times n}$ with $k \geq n$ is called the \emph{analysis operator}, and the \emph{analyzed vector} $\OP\mathbf{s} \in \R{k}$ is assumed to be sparse, where sparsity is again measured via an appropriate function $g$. In contrast to the synthesis model, where a signal is fully described by the non-zero elements of $\mathbf{x}$, in the analysis model the zero elements of the analyzed vector $\OP\mathbf{s}$ described the subspace containing the signal. To emphasize this difference, the term \emph{cosparsity} has been introduced in \cite{csp:nam:2011}, which simply counts the number of zero elements of $\OP\mathbf{s}$. As the sparsity in the synthesis model depends on the chosen dictionary, the cosparsity of an analyzed signal depends on the choice of the analysis operator $\OP$.

Different analysis operators proposed in the literature include the fused Lasso \cite{aop:tibshirani:2005}, the translation invariant wavelet transform \cite{aop:selesnick:2009}, and probably best known the finite difference operator which is closely related to the total-variation \cite{tv:rudin:1992}. They all have shown very good performance when used within the analysis model for solving diverse inverse problems in imaging. The question is: \emph{Can the performance of analysis based signal reconstruction be improved when a learned analysis operator is applied instead of a predefined one, as it is the case for the synthesis model where learned dictionaries outperform analytic dictionaries?}
In \cite{l1l2:elad:2007}, it has been discussed that the two models differ significantly, and the na\"{i}ve way of learning a dictionary and simply employing its transposed or its pseudo-inverse as the learned analysis operator fails. Hence, different algorithms are required for analysis operator learning.

\subsection{Contributions}
In this work, we introduce a new algorithm based on geometric optimization for learning a patch based analysis operator from a set of training samples, which we name GOAL (GeOmetric Analysis operator Learning). The method relies on a minimization problem, which is carefully motivated in Section \ref{sec:problem}. Therein, we also discuss the question of what is a suitable analysis operator for image reconstruction, and how to antagonize overfitting the operator to a subset of the training samples. An efficient geometric conjugate gradient method on the so-called oblique manifold is proposed in Section \ref{sec:manicg} for learning the analysis operator. Furthermore, in Section \ref{sec:image_rec} we explain how to apply the local patch based analysis operator to achieve global reconstruction results. Section \ref{sec:experments} sheds some light on the influence of the parameters required by GOAL and how to select them, and compares our method to other analysis operator learning techniques. The quality of the operator learned by GOAL on natural image patches is further investigated in terms of image denoising, inpainting, and single image super-resolution. The numerical results show the broad and effective applicability of our general approach.

\subsection{Notations}
Matrices are written as capital calligraphic letters like $\mathcal{X}$, column vectors are denoted by boldfaced small letters e.g.\ $\mathbf{x}$, whereas scalars are either capital or small letters like $n,N$. By $v_i$ we denote the $i^{\textit{th}}$ element of the vector $\mathbf{v}$, $v_{ij}$ denotes the $i^{\textit{th}}$ element in the $j^{\textit{th}}$ column of a matrix $\mathcal{V}$. The vector $\mathbf{v}_{:,i}$ denotes the $i^{\textit{th}}$ column of $\mathcal{V}$ whereas $\mathbf{v}_{i,:}$  denotes the transposed of the $i^{\textit{th}}$ row of $\mathcal{V}$. By $\mathcal{E}_{ij}$, we denote a matrix whose $i^{\textit{th}}$ entry in the $j^{\textit{th}}$ column is equal to one, and all others are zero. $\mathcal{I}_k$ denotes the identity matrix of dimension $(k \times k)$, ${\bm 0}$ denotes the zero-matrix of appropriate dimension, and $\operatorname{ddiag}(\mathcal{V})$ is the diagonal matrix whose entries on the diagonal are those of $\mathcal{V}$. By $\|\mathcal{V}\|_F^2=\sum_{i,j}v_{ij}^2$ we denote the squared Frobenius norm of a matrix $\mathcal{V}$, $\tr(\mathcal{V})$ is the trace of $\mathcal{V}$, and $\rk(\mathcal{V})$ denotes the rank.

\section{Analysis Operator Learning}

\subsection{Prior Art}\label{subsec:prior_art}
The topic of analysis operator learning has only recently started to be investigated, and only few prior work exists. In the sequel, we shortly review analysis operator learning methods that are applicable for image processing tasks.

Given a set of $M$ training samples $\big\{\mathbf{s}_i \in \R{n}\big\}_{i=1}^M$, the goal of analysis operator learning is to find a matrix $\OP \in \R{k \times n}$ with $k\geq n$, which leads to a maximally cosparse representation $\OP \mathbf{s}_i$ of each training sample. As mentioned in Subsection \ref{sec:syn_dict}, the training samples are distinctive vectorized image patches extracted from a set of example images. Let $\mathcal{S}=[\mathbf{s}_1,\dots,\mathbf{s}_M] \in \R{n \times M}$ be a matrix where the training samples constitute its columns, then the problem is to find
\begin{align}\label{eq:learnprob}
\OP^\star = \operatorname*{arg~min}_{\OP} \ G(\OP\mathcal{S}),
\end{align}
where $\OP$ is subject to some constraints, and $G$ is some function that measures the sparsity of the matrix $\OP \mathcal{S}$.
In \cite{aol:ophir:2011}, an algorithm is proposed in which the rows of the analysis operator are found sequentially by identifying directions that are orthogonal to a subset of the training samples. Starting from a randomly initialized vector ${\bm\omega} \in \R{n}$, a candidate row is found by first computing the inner product of ${\bm\omega}$ with the entire training set, followed by extracting the reduced training set $\mathcal{S}_R$ of samples whose inner product with ${\bm\omega}$ is smaller than a threshold. Thereafter, ${\bm\omega}$ is updated to be the eigenvector corresponding to the smallest eigenvalue of $\mathcal{S}_R\mathcal{S}_R^\top$. This procedure is iterated several times until a convergence criterion is met. If the determined candidate vector is sufficiently distinctive from already found ones, it is added to $\OP$ as a new row, otherwise it is discarded. This process is repeated until the desired number $k$ of rows have been found.

An adaption of the widely known K-SVD dictionary learning algorithm to the problem of analysis operator learning is presented in \cite{aol:rubinstein:2012}. As in the original K-SVD algorithm, $G(\OPo\mathcal{S}) = \sum_{i}\|\OPo\mathbf{s}_i\|_0$  is employed as the sparsifying function and the target cosparsity is required as an input to the algorithm. The arising optimization problem is solved by alternating between a sparse coding stage over each training sample while fixing $\OP$ using an ordinary analysis pursuit method, and updating the analysis operator using the optimized training set. Then, each row of $\OP$ is updated in a similar way as described in the previous paragraph for the method of \cite{aol:ophir:2011}. Interestingly, the operator learned on piecewise constant image patches by \cite{aol:ophir:2011} and \cite{aol:rubinstein:2012} closely mimics the finite difference operator.

In \cite{aol:yaghoobi:2011}, the authors use $G(\OPo\mathcal{S}) = \sum_{i}\|\OPo\mathbf{s}_i\|_1$ as the sparsity promoting function and suggest a constrained optimization technique that utilizes a projected subgradient method for iteratively solving \eqref{eq:learnprob}. To exclude the trivial solution, the set of possible analysis operators is restricted to the set of Uniform Normalized Tight Frames, i.e.\ matrices with uniform row norm and orthonormal columns. The authors state that this algorithm has the limitation of requiring noiseless training samples whose analyzed vectors $\{\OP \mathbf{s}_i \}_{i=1}^M$ are exactly cosparse.

To overcome this restriction, the same authors propose an extension of this algorithm that simultaneously learns the analysis operator and denoises the training samples, cf. \cite{aol:yaghoobi:2012}. This is achieved by alternating between updating the analysis operator via the projected subgradient algorithm and denoising the samples using an Augmented Lagrangian method. Therein, the authors state that their results for image denoising using the learned operator are only slightly worse compared to employing the commonly used finite difference operator.

An interesting idea related to the analysis model, called Fields-of-Experts (FoE) has been proposed in \cite{mrf:roth:2009}. The method relies on learning high-order Markov Random Field image priors with potential functions extending over large pixel neighborhoods, i.e. overlapping image patches. Motivated by a probabilistic model, they use the student-t distribution of several linear filter responses as the potential function, where the filters, which correspond to atoms from an analysis operator point of view, have been learned from training patches. Compared to our work and the methods explained above, their learned operator used in the experiments is underdetermined, i.e. $k<n$, the algorithms only works for small patches due to computational reasons, and the atoms are learned independently, while in contrast GOAL updates the analysis operator as a whole.

\subsection{Motivation of Our Approach}\label{sec:problem}
\begin{figure}
\centering
\includegraphics[width=\figwidth]{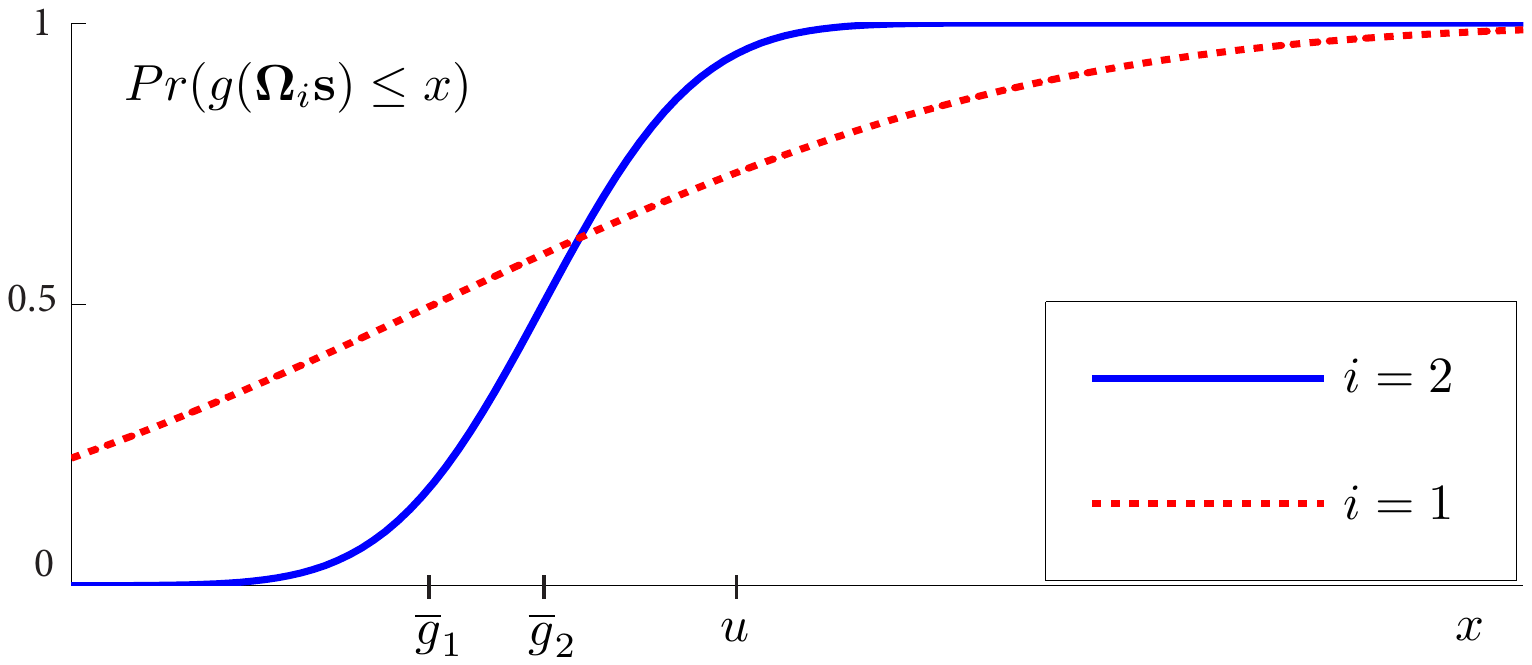}
\caption{Illustration of two possible distributions $Pr(g(\OP_i \Vector{s})\leq x)$ 
for two analysis operators. $\OP_1$: low expectation $\overline{g}_1$, high variance (dashed line); $\OP_2$: moderate expectation $\overline{g}_2$, moderate variance. Although $\OP_1$ yields a smaller expectation, there are more signals compared to $\OP_2$ where the sparsity model fails, i.e. $Pr(g(\OP_1 \Vector{s}) \geq u) > Pr(g(\OP_2 \Vector{s}) \geq u)$ for a suitable upper bound $u$.}
\label{fig:distributions} 
\end{figure}

%
%
In the quest for designing an analysis operator learning algorithm, the natural question arises: \emph{What is a good analysis operator for our needs?}
Clearly, given a signal $\mathbf{s}$ that belongs to a certain signal class, 
the aim is to find an $\OP$ such that $\OP \mathbf{s}$ is \emph{as sparse as possible}. This motivates to minimize the \emph{expected sparsity} $\mathbb{E}[g(\OP \mathbf{s})]$. All approaches presented in Subsection \ref{subsec:prior_art} can be explained in this way, i.e. for their sparsity measure $g$ they aim at learning an $\OP$ that minimizes the empirical mean of the sparsity over all randomly drawn training samples.
This, however, does not necessarily mean to learn \emph{the optimal} $\OP$ if the purpose is to reconstruct several signals belonging to a diverse class, e.g. natural image patches. The reason for this is that even if the \emph{expected} sparsity is low, 
it may happen with high probability that some realizations of this signal class cannot be represented in a sparse way, i.e. that for a given upper bound $u$, the probability $Pr(g(\OP \mathbf{s})\geq u)$ exceeds a tolerable value, cf. Figure \ref{fig:distributions}. 
%

The algorithm presented here aims at minimizing the empirical expectation of a sparsifying function $g(\OP \Vector{s}_i)$ for all training samples $\Vector{s}_i$, while additionally keeping the empirical variance 
moderate.
In other words, we try to avoid that the analyzed vectors of many similar training samples become \emph{very sparse} and consequently prevent $\OP$ from being adapted to the remaining ones. For image processing, this is of particular interest if the training patches are chosen randomly from natural images, because there is a high probability of collecting a large subset of very similar patches, e.g. homogeneous regions, that bias the learning process.

 Concretely, we want to find an $\OP$ that minimizes both the squared empirical mean 
\begin{align}
\overline{g}^2=\big(\tfrac{1}{M} \sum\limits_{i} g(\OP \Vector{s}_i) \big)^2
\end{align}
and the empirical variance 
\begin{align}
s^2=\tfrac{1}{M} \sum\limits_{i} \left(g(\OP \Vector{s}_i) - \overline{g} \right)^2
\end{align}
of the sparsity of the analyzed vectors. We achieve this by minimizing the sum of both, which is readily given by
\begin{align}
\overline{g}^2 + s^2 = \tfrac{1}{M} \sum\limits_{i} g(\OP \Vector{s}_i)^2.
\end{align}
Using $g(\cdot)=\|\cdot\|_p^p$, and introducing the factor $\tfrac{1}{2}$ the function we employ reads as
\begin{align}\label{eq:pqnorm}
J_{p}(\mathcal{V}) := \tfrac{1}{2 M}\sum\limits_{j=1}^M \big(\tfrac{1}{p}\sum\limits_{i=1}^k|v_{ij}|^p\big)^2 =  \tfrac{1}{2 M }\sum\limits_{j=1}^M\big(\tfrac{1}{p}\|\mathbf{v}_{:,j}\|_p^p\big)^2,
\end{align}
with $0 \leq p \leq 1$ and $\mathcal{V}=\OP \mathcal{S}$. 

Certainly, without additional prior assumptions on $\OP$, the useless solution $\OP =\mathbf{0}$ is the global minimizer of Problem \eqref{eq:learnprob}. To avoid the trivial solution and for other reasons explained later in this section, we regularize the problem by imposing the following three constraints on $\OP$.
\begin{enumerate}[(i)]
\item The \emph{rows} of $\OP$ have unit Euclidean norm, i.e.\ $\|{\bm\omega}_{i,:}\|_2=1$ for $i=1,\ldots,k$.
\item The analysis operator $\OP$ has full rank, i.e.\ $\rk(\OP)=n$.
\item The analysis operator $\OP$ does not have linear dependent rows, i.e.\ ${\bm\omega}_{i,:} \neq \pm {\bm\omega}_{j,:}$ for $i \neq j$.
\end{enumerate}
The rank condition (ii) on $\OP$ is motivated by the fact that different input samples $\mathbf{s}_1,\mathbf{s}_2 \in \R{n}$ with  $\mathbf{s}_1 \neq \mathbf{s}_2$ should be mapped to different analyzed vectors $\OP\mathbf{s}_1 \neq \OP\mathbf{s}_2$. With Condition (iii) redundant transform coefficients in an analyzed vector are avoided.

These constraints motivate the consideration of the set of full rank matrices with normalized \emph{columns}, which admits a manifold structure known as the \emph{oblique manifold }\cite{ob:trandafilov:1999}
\begin{align}\label{eq:ob_man}
\mathrm{OB}(n,k) := \{\mathcal{X} \in \R{n \times k} | \rk(\mathcal{X})=n, \operatorname*{ddiag}(\mathcal{X}^\top \mathcal{X})=\mathcal{I}_k\}.
\end{align}
Note, that this definition only yields a non-empty set if $k \geq n$, which is the interesting case in this work. Thus, from now on, we assume $k \geq n$. Remember that we require the \emph{rows} of $\OP$ to have unit Euclidean norm. Hence, we restrict the \emph{transposed} of the learned analysis operator to be an element of $ \mathrm{OB}(n,k)$.

Since $\mathrm{OB}(n,k)$ is open and dense in the set of matrices with normalized columns, we need a penalty function that ensures the rank constraint (ii) and prevents iterates to approach the boundary of $\mathrm{OB}(n,k)$.

\begin{lemma}\label{lemma:det}
The inequality $ 0 < \det(\tfrac{1}{k}\mathcal{X} \mathcal{X}^\top) \leq (\tfrac{1}{n})^n$ holds true for all $\mathcal{X} \in \mathrm{OB}(n,k)$, where $1 < n \leq k$.
\end{lemma}

\begin{IEEEproof}
Due to the full rank condition on $\mathcal{X}$, the product $\mathcal{X} \mathcal{X}^\top$ is positive definite, consequently the strict inequality $0 < \det(\tfrac{1}{k}\mathcal{X} \mathcal{X}^\top)$ applies. To see the second inequality of Lemma \ref{lemma:det}, observe that
\begin{align}
\|\mathcal{X}\|_F^2= \tr (\mathcal{X} \mathcal{X}^\top)=k,
\end{align}
which implies $\tr (\tfrac{1}{k} \mathcal{X} \mathcal{X}^\top) = 1$. Since the trace of a matrix is equal to the sum of its eigenvalues, which are strictly positive in our case, it follows that the strict inequality $0<\lambda_i<1$ holds true for all eigenvalues $\lambda_i$ of $\tfrac{1}{k} \mathcal{X} \mathcal{X}^\top$. From the well known relation between the arithmetic and the geometric mean we see
\begin{align}\label{eq:arithmean}
\sqrt[n]{\Pi \lambda_i } \leq \tfrac{1}{n} \sum \lambda_i.
\end{align}
Now, since the determinant of a matrix is equal to the product of its eigenvalues,
and with $\sum \lambda_i = \tr (\tfrac{1}{k} \mathcal{X} \mathcal{X}^\top) = 1$, we have
\begin{equation}\label{eq:afterarithmeticmean}
\det(\tfrac{1}{k}\mathcal{X} \mathcal{X}^\top)= \Pi \lambda_i \leq (\tfrac{1}{n})^n,
\end{equation}
which completes the proof.
\end{IEEEproof}

Recalling that $\OP^\top \in \mathrm{OB}(n,k)$ and considering Lemma \ref{lemma:det}, we can enforce the full rank constraint with the penalty function
\begin{align}\label{eq:logdet}
h(\OP):= -\tfrac{1}{n\log(n)} \log \det(\tfrac{1}{k} \OP^\top \OP).
\end{align}

Regarding Condition (iii), the following result proves useful.
\begin{lemma}\label{lemma:rnorm}
For a matrix $\mathcal{X} \in \mathrm{OB}(n,k)$ with $1 < n \leq k$, 
the inequality $|\mathbf{x}_{:,i}^\top \mathbf{x}_{:,j}| \leq 1$ applies, where equality holds true if and only if $\mathbf{x}_{:,i}=\pm \mathbf{x}_{:,j}$.
\end{lemma}

\begin{IEEEproof}
By the definition of $\mathrm{OB}(n,k)$ the columns of $\mathcal{X}$ are normalized, consequently Lemma \ref{lemma:rnorm} follows directly from the Cauchy-Schwarz inequality.
\end{IEEEproof}

Thus, Condition (iii) can be enforced via the logarithmic barrier function of the scalar products between all distinctive rows of $\OP$, i.e.\
\begin{align}\label{eq:lindep}
r(\OP):= - \hspace{-4mm} \sum \limits_{1 \leq i < j \leq k} \log(1-({\bm \omega}_{i,:}^\top {\bm \omega}_{j,:})^2).
\end{align}

Finally, combining all the introduced constraints, our optimization problem for learning the transposed analysis operator reads as
\begin{align}\label{eq:opt_prob}
\OP^{\top}  = \operatorname*{arg~min}_{\mathcal{X}\in \mathrm{OB}(n,k)} J_{p}(\mathcal{X}^\top \mathcal{S}) + \kappa \ h(\mathcal{X}^\top) + \mu \ r(\mathcal{X}^\top).
\end{align}
Therein, the two weighting factors $\kappa,\mu \in \R{+}$ control the influence of the two constraints on the final solution.
The following lemma clarifies the role of $\kappa$.
\begin{lemma}\label{lemma:cond}
Let $\OP$ be a minimum of $h$ in the set of transposed oblique matrices, i.e.
\begin{align}
\OP^\top \in \underset{\mathcal{X}  \in \mathrm{OB}(n,k)}{\operatorname{arg~min}} h(\mathcal{X}^\top),
\end{align}
then the condition number of $\OP$ is equal to one.
\end{lemma}
\begin{IEEEproof}
It is well known that equality of the arithmetic and the geometric mean in Equation \eqref{eq:arithmean} holds true, if and only if all eigenvalues $\lambda_i$ of $\tfrac{1}{k} \mathcal{X} \mathcal{X}^\top$ are equal, i.e.\
$\lambda_1 = \ldots = \lambda_n$. Hence, if $\OP^\top \in \underset{\mathcal{X}  \in \mathrm{OB}(n,k)}{\operatorname{arg~min}} h(\mathcal{X}^\top)$, then $\det(\tfrac{1}{k} \OP^\top \OP)= (\tfrac{1}{n})^n$, and consequently all singular values of $\OP$ coincide. This implies that the condition number of $\OP$, which is defined as the quotient of the largest to the smallest singular value, is equal to one.
\end{IEEEproof}
With other words, the minima of $h$ are uniformly normalized tight frames, cf.  \cite{aol:yaghoobi:2011,aol:yaghoobi:2012}.
From Lemma \ref{lemma:cond} we can conclude that with larger $\kappa$ the condition number of $\OP$ approaches one.
Now, recall the inequality
\begin{align}\label{eq:dist}
\sigma_{\min} \|\mathbf{s}_1-\mathbf{s}_2\|_2 \leq \| \OP(\mathbf{s}_1-\mathbf{s}_2) \|_2 \leq \sigma_{\max} \|\mathbf{s}_1-\mathbf{s}_2\|_2,
\end{align}
with $\sigma_{\min}$ being the smallest and $\sigma_{\max}$ being the largest singular value of $\OP$. From this it follows that an analysis operator found with a large $\kappa$, i.e.\ obeying $\sigma_{\min} \approx \sigma_{\max}$, carries over distinctness of different signals to their analyzed versions. 
%
The parameter $\mu$ regulates the redundancy between the rows of the analysis operator and consequently avoids redundant coefficients in the analyzed vector $\OP \mathbf{s}$.
\begin{lemma}\label{lemma:mc}
The difference between any two entries of the analyzed vector $\OP \mathbf{s}$ is bounded by
\begin{align}
|\bm{\omega}_{i,:}^\top\mathbf{s} - \bm{\omega}_{j,:}^\top\mathbf{s}| \leq \sqrt{2 (1- \bm{\omega}_{i,:}^\top\bm{\omega}_{j,:})} \ \| \mathbf{s} \|_2.
\end{align}
\end{lemma}
\begin{IEEEproof}
From the Cauchy-Schwarz inequality we get
\begin{align}
|\bm{\omega}_{i,:}^\top\mathbf{s} - \bm{\omega}_{j,:}^\top\mathbf{s}| = | (\bm{\omega}_{i,:}- \bm{\omega}_{j,:})^\top \mathbf{s} | \leq \|\bm{\omega}_{i,:}- \bm{\omega}_{j,:} \|_2 \|\mathbf{s}\|_2.
\end{align}
Since by definition $\| \bm{\omega}_{i,:} \|_2 = \| \bm{\omega}_{j,:} \|_2 = 1$, it follows that $\|\bm{\omega}_{i,:}- \bm{\omega}_{j,:} \|_2= \sqrt{2(1- \bm{\omega}_{i,:}^\top \bm{\omega}_{j,:})}$.
\end{IEEEproof}

The above lemma implies, that if the $i^\textit{th}$ entry of the analyzed vector is significantly larger than $0$ then a large absolute value of $\bm{\omega}_{i,:}^\top\bm{\omega}_{j,:}$ prevents
 the $j^\textit{th}$ entry to be small. 
To achieve large cosparsity, this is an unwanted effect that our approach avoids via the log-barrier function $r$ in \eqref{eq:lindep}. It is worth mentioning that the same effect is achieved by minimizing the analysis operator's mutual coherence $\max\limits_{i \neq j}|{\bm \omega}_{i,:}^\top {\bm \omega}_{j,:}|$ and that 
 our experiments suggest that enlarging $\mu$ leads to minimizing the mutual coherence. 

In the next section, we explain how the manifold structure of $\mathrm{OB}(n,k)$ can be exploited to efficiently learn the analysis operator.

\section{Analysis Operator Learning Algorithm}\label{sec:manicg}
Knowing that the feasible set of solutions to Problem \eqref{eq:opt_prob} is restricted to a smooth manifold allows us to formulate a geometric conjugate gradient (CG-) method to learn the analysis operator. Geometric CG-methods have been proven efficient in various applications, due to the combination of moderate computational complexity and good convergence properties, see e.g.\ \cite{cg:junior:2012} for a CG-type method on the oblique manifold.

To make this work self contained, we start by shortly reviewing the general concepts of optimization on matrix manifolds. After that we present the concrete formulas and implementation details for our optimization problem on the oblique manifold. For an in-depth introduction on optimization on matrix manifolds, we refer the interested reader to \cite{mani:absil:2008}.
\begin{figure}
\centering
\includegraphics[width=\figwidth]{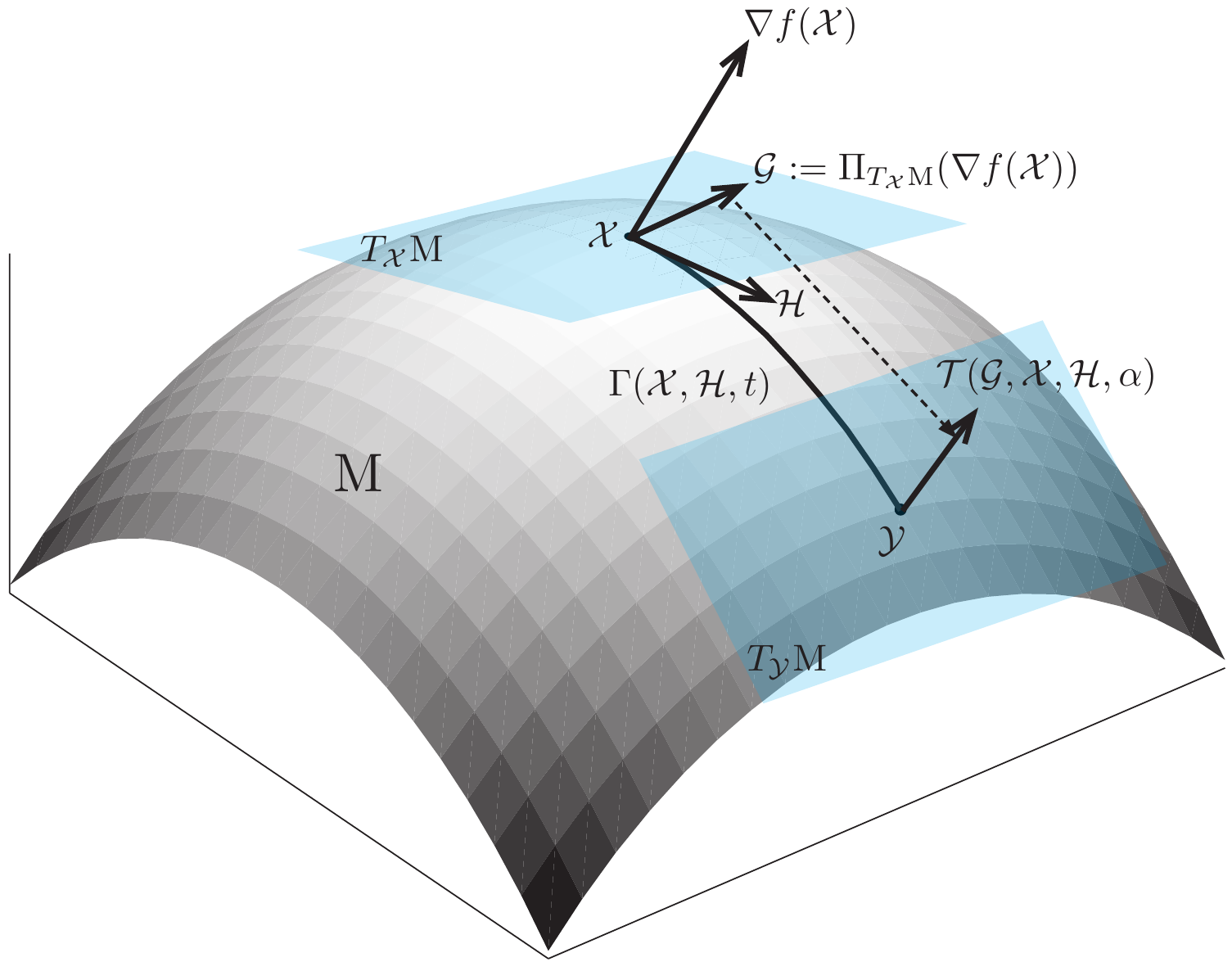}
\caption{This figure shows two points $\mathcal{X}$ and $\mathcal{Y}$ on a manifold $\mathrm{M}$ together with their tangent spaces $T_\mathcal{X}\mathrm{M}$ and $T_\mathcal{Y}\mathrm{M}$. Furthermore, the Euclidean gradient $\nabla f(\mathcal{X})$ and its projection onto the tangent space $\Pi_{T_\mathcal{X}\mathrm{M}}(\nabla f(\mathcal{X}))$ are depicted. The geodesic $\Gamma(\mathcal{X},\mathcal{H},t)$ in the direction of $\mathcal{H} \in T_\mathcal{X}\mathrm{M}$ connecting the two points is shown. The dashed line typifies the role of a parallel transport of the gradient in $T_\mathcal{X}\mathrm{M}$ to $T_\mathcal{Y}\mathrm{M}$.} \label{fig:manifold} \end{figure}

\subsection{Optimization on Matrix Manifolds}
Let $\mathrm{M}$ be a smooth Riemannian submanifold of $\R{n \times k}$ with the standard Frobenius inner product $\langle \mathcal{Q} ,\mathcal{P} \rangle := \tr (\mathcal{Q}^\top \mathcal{P})$, and let $f \colon \R{n \times k} \to \R{}$ be a differentiable cost function.
We consider the problem of finding
\begin{align}\label{eq:optmani}
\operatorname*{arg~min}_{\mathcal{X}\in \mathrm{M}} f(\mathcal{X}).
\end{align}
The concepts presented in this subsection are visualized in Figure \ref{fig:manifold} to alleviate the understanding.

To every point $\mathcal{X} \in \mathrm{M}$ one can assign a tangent space $T_\mathcal{X}\mathrm{M}$. The tangent space at $\mathcal{X}$ is a real vector space containing all possible directions that tangentially pass through $\mathcal{X}$. An element $\Xi \in T_\mathcal{X}\mathrm{M}$ is called a tangent vector at $\mathcal{X}$. Each tangent space is associated with an inner product inherited from the surrounding $\R{n \times k}$, which allows to measure distances and angles on $\mathrm{M}$.

The Riemannian gradient of $f$ at $\mathcal{X}$ is an element of the tangent space $T_\mathcal{X}\mathrm{M}$ that points in the direction of steepest ascent of the cost function on the manifold. As we require $\mathrm{M}$ to be a submanifold of $\R{n \times k}$ and since by assumption $f$ is defined on the whole $\R{n \times k}$, the Riemannian gradient $\mathcal{G}(\mathcal{X})$ is simply the orthogonal projection of the (standard) gradient $\nabla f(\mathcal{X})$ onto the tangent space $T_\mathcal{X}\mathrm{M}$. In formulas, this reads as
\begin{align}\label{eq:rgrad}
\mathcal{G}(\mathcal{X}):=\Pi_{T_\mathcal{X}\mathrm{M}}(\nabla f(\mathcal{X})).
\end{align}

A \emph{geodesic} is a smooth curve $\Gamma(\mathcal{X},\Xi,t)$ emanating from $\mathcal{X}$ in the direction of $\Xi \in T_\mathcal{X}\mathrm{M}$, which locally describes the shortest path between two points on $\mathrm{M}$. Intuitively, it can be interpreted as the equivalent of a straight line in the manifold setting.

Conventional line search methods search for the next iterate along a straight line. This is generalized to the manifold setting as follows. Given a current optimal point $\mathcal{X}^{(i)}$ and a search direction $\mathcal{H}^{(i)} \in T_{\mathcal{X}^{(i)}}\mathrm{M}$ at the $i^\textit{th}$ iteration, the step size $\alpha^{(i)}$ which leads to sufficient decrease of $f$ can be determined by finding the minimizer of
\begin{align}\label{eq:linesearch}
\alpha^{(i)} = \operatorname*{arg~min}_{t \geq 0} f(\Gamma(\mathcal{X}^{(i)},\mathcal{H}^{(i)}, t)).
\end{align}
Once $\alpha^{(i)}$ has been determined, the new iterate is computed by
\begin{align}\label{eq:update}
\mathcal{X}^{(i+1)} = \Gamma(\mathcal{X}^{(i)},\mathcal{H}^{(i)}, \alpha^{(i)}).
\end{align}

Now, one straightforward approach to minimize $f$ is to alternate Equations \eqref{eq:rgrad}, \eqref{eq:linesearch}, and \eqref{eq:update} using $\mathcal{H}^{(i)}=-\mathcal{G}^{(i)}$, with the short hand notation
$\mathcal{G}^{(i)}:=\mathcal{G}(\mathcal{X}^{(i)})$, which corresponds to the steepest descent on a Riemannian manifold. However, as in standard optimization, steepest descent only has a linear rate of convergence. Therefore, we employ a conjugate gradient method on a manifold, as it offers a superlinear rate of convergence, while still being applicable to large scale optimization problems with low computational complexity.

In CG-methods, the updated search direction $\mathcal{H}^{(i+1)} \in T_{\mathcal{X}^{(i+1)}}\mathrm{M}$ is a linear combination of the gradient $\mathcal{G}^{(i+1)}\in T_{\mathcal{X}^{(i+1)}}\mathrm{M}$ and the previous search direction $\mathcal{H}^{(i)} \in T_{\mathcal{X}^{(i)}}\mathrm{M}$. Since adding vectors that belong to different tangent spaces is not defined, we need to map $\mathcal{H}^{(i)}$ from $T_{\mathcal{X}^{(i)}}\mathrm{M}$ to $T_{\mathcal{X}^{(i+1)}}\mathrm{M}$. This is done by the so-called parallel transport $\mathcal{T}(\Xi,\mathcal{X}^{(i)},\mathcal{H}^{(i)},\alpha^{(i)})$, which transports a tangent vector $\Xi \in T_{\mathcal{X}^{(i)}}\mathrm{M}$ along the geodesic $\Gamma(\mathcal{X}^{(i)},\mathcal{H}^{(i)},t)$ to the tangent space $T_{\mathcal{X}^{(i+1)}}\mathrm{M}$. Now, using the shorthand notation
\begin{align}\label{eq:shorthand}
\mathcal{T}^{(i+1)}_{\Xi}:=\mathcal{T}(\Xi,\mathcal{X}^{(i)},\mathcal{H}^{(i)},\alpha^{(i)}),
\end{align}
the new search direction is computed by
\begin{align}\label{eq:sdir}
\mathcal{H}^{(i+1)} = -\mathcal{G}^{(i+1)} + \beta^{(i)}\mathcal{T}^{(i+1)}_{\mathcal{H}^{(i)}},
\end{align}
where $\beta^{(i)} \in \R{}$ is calculated by some update formula adopted to the manifold setting.
Most popular are the update formulas by Fletcher-Reeves (FR), Hestenes-Stiefel (HS), and Dai-Yuan (DY).
With $\mathcal{Y}^{(i+1)} = \mathcal{G}^{(i+1)}-\mathcal{T}^{(i+1)}_{\mathcal{G}^{(i)}}$, they read as 
\begin{align}
\beta^{(i)}_{\textit{FR}} & =  \frac{\langle \mathcal{G}^{(i+1)},\mathcal{G}^{(i+1)}\rangle}{\langle \mathcal{G}^{(i)},\mathcal{G}^{(i)}\rangle},\label{eq:fr}\\
\beta^{(i)}_{\textit{HS}} & =  \frac{\langle\mathcal{G}^{(i+1)},\mathcal{Y}^{(i+1)} \rangle}{\langle \mathcal{T}^{(i+1)}_{\mathcal{H}^{(i)}},\mathcal{Y}^{(i+1)} \rangle},\label{eq:hs}\\
\beta^{(i)}_{\textit{DY}} & =  \frac{\langle \mathcal{G}^{(i+1)},\mathcal{G}^{(i+1)}\rangle}{\langle \mathcal{T}^{(i+1)}_{\mathcal{H}^{(i)}},\mathcal{Y}^{(i+1)} \rangle}\label{eq:dy}.
\end{align}

Now, a solution to Problem \eqref{eq:optmani} is computed by alternating between finding the search direction on $\mathrm{M}$ and updating the current optimal point until some user-specified convergence criterion is met, or a maximum number of iterations has been reached.

\subsection{Geometric Conjugate Gradient for Analysis Operator Learning}
In this subsection we derive all ingredients to implement the geometric conjugate gradient method as described in the previous subsection for the task of learning the analysis operator. Results regarding the geometry of $\mathrm{OB}(n,k)$ are derived e.g.\ in \cite{mani:absil:2008}. To enhance legibility, and since the dimensions $n$ and $k$ are fixed throughout the rest of the paper, the oblique manifold is further on denoted by $\mathrm{OB}$.

The tangent space at $\mathcal{X} \in  \mathrm{OB}$ is given by
\begin{align}\label{eq:t_space}
T_\mathcal{X}\mathrm{OB} = \{\Xi \in \R{n\times k}|\operatorname{ddiag}(\mathcal{X}^\top\Xi) = {\bm 0}\}.
\end{align}
The orthogonal projection of a matrix $\mathcal{Q} \in \R{n \times k}$ onto the tangent space $T_\mathcal{X}\mathrm{OB}$ is
\begin{align}\label{eq:oproj}
\Pi_{T_\mathcal{X}\mathrm{OB}}(\mathcal{Q}) = \mathcal{Q} - \mathcal{X}\operatorname{ddiag}(\mathcal{X}^\top\mathcal{Q}).
\end{align}
Regarding geodesics, note that in general a geodesic is the solution of a second order ordinary differential equation, meaning that for arbitrary manifolds, its computation as well as computing the parallel transport is not feasible. Fortunately, as the oblique manifold is a Riemannian submanifold of a product of $k$ unit spheres $S^{n-1}$, the formulas for parallel transport and the exponential mapping allow an efficient implementation.

Let $\mathbf{x} \in S^{n-1}$ be a point on a sphere and $\mathbf{h} \in T_\mathbf{x}S^{n-1}$ be a tangent vector at $\mathbf{x}$, then the geodesic in the direction of $\mathbf{h}$ is a great circle \begin{align}\label{eq:mapping}
\gamma(\mathbf{x},\mathbf{h}, t)&=
\left\{
\begin{array}{lr}
\mathbf{x}, & \textit{if } \|\mathbf{h}\|_2= 0\\
\mathbf{x}\cos(t \|\mathbf{h}\|_2)+\mathbf{h} \frac{\sin(t \|\mathbf{h}\|_2)}{\|\mathbf{h}\|_2}, & \textit{otherwise.}
\end{array} \right.
\end{align}
The associated parallel transport of a tangent vector $\bm{\xi} \in T_\mathbf{x}S^{n-1}$ along the great circle $\gamma(\mathbf{x},\mathbf{h},t)$ reads as
\begin{align}\label{eq:patrus}
\tau(\bm{\xi},\mathbf{x},\mathbf{h},t) =\bm{\xi}-\frac{\bm{\xi}^\top \mathbf{h}}{\|\mathbf{h}\|_2^2}\bigg( & \mathbf{x}\|\mathbf{h}\|_2\sin(t\|\mathbf{h}\|_2)+ \nonumber \\ & \mathbf{h}(1-\cos(t\|\mathbf{h}\|_2))\bigg).
\end{align}


As $\mathrm{OB}$ is a submanifold of the product of unit spheres, the geodesic through $\mathcal{X} \in \mathrm{OB}$ in the direction of $\mathcal{H} \in T_\mathcal{X}\mathrm{OB}$ is simply the combination of the great circles emerging by concatenating each column of $\mathcal{X}$ with the corresponding column of $\mathcal{H}$, i.e.\
\begin{align}\label{eq:mapping_mtx}
\Gamma(\mathcal{X},\mathcal{H}, t) = \bigg[\gamma(\mathbf{x}_{:,1},\mathbf{h}_{:,1},t),\ldots,\gamma(\mathbf{x}_{:,k},\mathbf{h}_{:,k},t)\bigg].
\end{align}
Accordingly, the parallel transport of $\Xi \in T_\mathcal{X}\mathrm{OB}$ along the geodesic $\Gamma(\mathcal{X},\mathcal{H},t)$ is given by
\begin{align}\label{eq:patrus_mtx}
\begin{split}
& \mathcal{T}(\Xi,\mathcal{X},\mathcal{H},t) = \\ &  \quad \bigg[\tau(\bm{\xi}_{:,1},\mathbf{x}_{:,1},\mathbf{h}_{:,1},t),\ldots,\tau(\bm{\xi}_{:,k},\mathbf{x}_{:,k},\mathbf{h}_{:,k},t)\bigg].
\end{split}
\end{align}

Now, to use the geometric CG-method for learning the analysis operator, we require a differentiable cost function $f$. Since, the cost function presented in Problem \eqref{eq:opt_prob} is not differentiable due to the non-smoothness of the $(p,q)$-pseudo-norm \eqref{eq:pqnorm},
we exchange Function \eqref{eq:pqnorm} with a smooth approximation, which is given by
\begin{align}\label{eq:pq_sm}
J_{p,\nu}(\mathcal{V}) := \tfrac{1}{2 M}\sum\limits_{j=1}^M \left(\tfrac{1}{p}\sum\limits_{i=1}^k(v_{ij}^2+\nu)^\frac{p}{2}\right)^2,
\end{align}
with $\nu \in \R{+}$ being the smoothing parameter. The smaller $\nu$ is, the more closely the approximation resembles the original function.
Again, taking $\mathcal{V}=\Omega \mathcal{S}$ and with the shorthand notation $z_{ij}:=(\Omega \mathcal{S})_{ij}$, the gradient of the applied sparsity promoting function \eqref{eq:pq_sm} reads as
\begin{align}\label{eq:gradg}
\tfrac{\partial}{\partial \OP} J_{p,\nu}(\OP\mathcal{S}) = & \left[\tfrac{1}{M}
\sum\limits_{j=1}^M \tfrac{1}{p}\sum\limits_{i=1}^k(z_{ij}^2+\nu)^\frac{p}{2} \right. \nonumber\\ & \left.\sum\limits_{i=1}^k\left\{z_{ij}(z_{ij}^2+\nu)^{\frac{p}{2}-1}\mathcal{E}_{ij}\right\}\right]\mathcal{S}^\top.
\end{align}
The gradient of the rank penalty term \eqref{eq:logdet} is
\begin{align}\label{eq:gradh}
\tfrac{\partial}{\partial \OP} h(\OP)= -\tfrac{2}{k n \log(n)} \OP (\tfrac{1}{k}\OP^\top \OP)^{-1}
\end{align}
and the gradient of the logarithmic barrier function \eqref{eq:lindep} is
\begin{align}\label{eq:gradr}
\tfrac{\partial}{\partial \OP} r(\OP)= \left[\sum \limits_{1 \leq i < j \leq k} \frac{2 {\bm \omega}_{i,:}^\top {\bm \omega}_{j,:}}{1-({\bm \omega}_{i,:}^\top {\bm \omega}_{j,:})^2}(\mathcal{E}_{ij}+\mathcal{E}_{ji})\right]\OP.
\end{align}

Combining Equations~\eqref{eq:gradg}, \eqref{eq:gradh}, and \eqref{eq:gradr}, the gradient of the cost function
\begin{align}\label{eq:opt_costfunc_smooth}
f(\mathcal{X}) := J_{p,\nu}(\mathcal{X}^\top \mathcal{S}) + \kappa \ h(\mathcal{X}^\top) + \mu \ r(\mathcal{X}^\top)
\end{align}
which is used for learning the analysis operator reads as
\begin{align}\label{eq:grad_costfunc}
\nabla f(\mathcal{X})= \tfrac{\partial}{\partial \mathcal{X}} J_{p,\nu}(\mathcal{X}^\top\mathcal{S}) + \kappa \tfrac{\partial}{\partial \mathcal{X}} h(\mathcal{X}^\top) + \mu \tfrac{\partial}{\partial \mathcal{X}} r(\mathcal{X}^\top).
\end{align}

Regarding the CG-update parameter $\beta^{(i)}$, we employ a hybridization of the Hestenes-Stiefel Formula \eqref{eq:hs} and the Dai Yuan formula \eqref{eq:dy}
\begin{align}\label{eq:dyhs}
\beta^{(i)}_{\textit{hyb}} = \max\big (0,\min(\beta^{(i)}_{\textit{DY}},\beta^{(i)}_{\textit{HS}})\big),
\end{align}
which has been suggested in \cite{cg:dai:2001}. As explained therein, formula \eqref{eq:dyhs} combines the good numerical performance of HS with the desirable global convergence properties of DY.

Finally, to compute the step size $\alpha^{(i)}$, we use an adaption of the well-known backtracking line search to the geodesic $\Gamma(\mathcal{X}^{(i)},\mathcal{H}^{(i)},t)$. In that, an initial step size $t^{(i)}_0$  is iteratively decreased by a constant factor $c_1 < 1$ until  the Armijo condition is met, see Algorithm \ref{al:backtrack} for the entire procedure.
\begin{algorithm}
\caption{Backtracking Line Search on Oblique Manifold} \label{al:backtrack}
\begin{algorithmic}
\STATE \hspace{-4.4mm} \textbf{Input:} $t_0^{(i)} > 0,\; 0 < c_1 < 1,\; 0 < c_2 < 0.5$, $\;\mathcal{X}^{(i)},\mathcal{G}^{(i)},\mathcal{H}^{(i)}$
\STATE \hspace{-4.4mm} \textbf{Set:} $t \leftarrow t_0^{(i)} $
\WHILE{$f(\Gamma(\mathcal{X}^{(i)},\mathcal{H}^{(i)}, t)) > f(\mathcal{X}^{(i)}) + t c_2 \langle \mathcal{G}^{(i)},\mathcal{H}^{(i)}\rangle$}
\STATE $t \leftarrow c_1 t$
\ENDWHILE
\STATE \hspace{-4.4mm} \textbf{Output:} $\alpha^{(i)} \leftarrow t $
\end{algorithmic}
\end{algorithm}
In our implementation we empirically chose $c_1 = 0.9$ and $c_2 = 10^{-2}$. As an initial guess for the step size at the first CG-iteration $i=0$, we choose
\begin{align}\label{eq:step_0}
t^{(0)}_0 =\|\mathcal{G}^{(0)}\|_F^{-1},
\end{align}
as proposed in \cite{cg:gilbert:1992}. In the subsequent iterations, the backtracking line search is initialized by the previous step size divided by the line search parameter, i.e.\ $t^{(i)}_0 = \frac{\alpha^{(i-1)}}{c_1}$. Our complete approach for learning the analysis operator is summarized in Algorithm \ref{al:learning}.
Note, that under the conditions that the Fletcher-Reeves update formula is used and some mild conditions on the step-size selection, the convergence of Algorithm \ref{al:learning} to a critical point, i.e.\ $\liminf_{i\to\infty}  \| \mathcal{G}^{(i)} \| = 0$, is guaranteed by a result provided in \cite{cg:ring:2012}.
\begin{algorithm}
\caption{Geometric Analysis Operator Learning (GOAL)}
\label{al:learning}
\begin{algorithmic}
\STATE \hspace{-4.4mm} \textbf{Input:} Initial analysis operator $\OP_{\textit{init}}$, training data $\mathcal{S}$, parameters $p,\nu,\kappa,\mu$
\STATE \hspace{-4.4mm} \textbf{Set:}  $i \leftarrow 0$, $\mathcal{X}^{(0)} \leftarrow \OP_{\textit{init}}^\top$, $\mathcal{H}^{(0)}\leftarrow-\mathcal{G}^{(0)}$
\REPEAT
\STATE $\alpha^{(i)} \leftarrow \underset{t \geq 0}{\arg \ \min} \ f(\Gamma(\mathcal{X}^{(i)},\mathcal{H}^{(i)}, t))$, cf. Algorithm \ref{al:backtrack} in conjunction with Equation \eqref{eq:opt_costfunc_smooth}\vspace{1mm}
\STATE $\mathcal{X}^{(i+1)} \leftarrow \Gamma(\mathcal{X}^{(i)},\mathcal{H}^{(i)}, \alpha^{(i)})$, cf. Equation \eqref{eq:mapping_mtx}\vspace{1mm}
\STATE $\mathcal{G}^{(i+1)} \leftarrow \Pi_{T_{\mathcal{X}^{(i+1)}}\mathrm{M}}(\nabla f(\mathcal{X}^{(i+1)}))$, cf. Equations \eqref{eq:oproj} and \eqref{eq:grad_costfunc} \vspace{1mm}
\STATE $\beta^{(i)}\leftarrow \max\big(0,\min(\beta^{(i)}_{\textit{DY}},\beta^{(i)}_{\textit{HS}})\big)$, cf. Equations \eqref{eq:hs}, \eqref{eq:dy}\vspace{1mm}
\STATE $\mathcal{H}^{(i+1)} \leftarrow -\mathcal{G}^{(i+1)} + \beta^{(i)}\mathcal{T}^{(i+1)}_{\mathcal{H}^{(i)}}$, cf. Equations \eqref{eq:shorthand}, \eqref{eq:patrus_mtx}\vspace{1mm}
\STATE $i \leftarrow i+1$
\UNTIL{$\|\mathcal{X}^{(i)}-\mathcal{X}^{(i-1)}\|_F<10^{-4}$ $\lor \ i = $ maximum $\#$ iterations}
\STATE \hspace{-4.4mm} \textbf{Output:} $\OP^\star \leftarrow \mathcal{X}^{(i)\top} $
\end{algorithmic}
\end{algorithm}

\section{Analysis Operator based Image Reconstruction}\label{sec:image_rec}
In this section we explain how the analysis operator $\OP^\star \in \R{k \times n}$ is utilized for reconstructing an unknown image $\mathbf{s} \in \R{N}$ from some measurements $\mathbf{y} \in \R{m}$ following the analysis approach \eqref{eq:std_min_an}. Here, the vector $\mathbf{s} \in \R{N}$ denotes a vectorized image of dimension $N=wh$, with $w$ being the width and $h$ being the height of the image, respectively, obtained by stacking the columns of the image above each other. In the following, we will loosely speak of $\mathbf{s}$ as the image.

Remember, that the size of $\OP^\star$ is very small compared to the size of the image, and it has to be applied locally to small image patches rather than globally to the entire image.
%
%
%
Artifacts that arise from na{\"i}ve patch-wise reconstruction are commonly reduced by considering overlapping patches. Thereby, each patch is reconstructed individually and the entire image is formed by averaging over the overlapping regions in a final step. 
However, this method misses global support during the reconstruction process, hence, it leads to poor inpainting results and is not applicable for e.g. Compressive Sensing tasks. To overcome these drawbacks, we use a method related to the patch based synthesis approach from \cite{dl:elad:2006} and the method used in \cite{mrf:roth:2009}, which provides global support from local information. Instead of optimizing over each patch individually and combining them in a final step, 
we optimize over the \emph{entire} image demanding that
a pixel is reconstructed such that 
the average sparsity of all patches it belongs to is minimized.
%
%
%
 When all possible patch positions are taken into account, this procedure is entirely partitioning-invariant. For legibility, we assume square patches i.e. of size $(\sqrt{n} \times \sqrt{n}$) with $\sqrt{n}$ being a positive integer. 

Formally, let $\mathbf{r} \subseteq \{1,\dots,h\}$ and $\mathbf{c} \subseteq \{1,\dots,w\}$ denote sets of indices 
with $r_{i+1}-r_{i} = d_v$, $c_{i+1}-c_{i} = d_h$ and $1 \leq d_v,d_h \leq \sqrt{n}$. Therein, $d_v,d_h$ determine the degree of overlap between two adjacent patches in vertical, and horizontal direction, respectively. We consider all image patches whose center is an element of the cartesian product set $\mathbf{r} \times \mathbf{c}$. Hence, with $|\cdot|$ denoting the cardinality of a set, the total number of patches being considered is equal to $|\mathbf{r}||\mathbf{c}|$.  Now, let $\mathcal{P}_{rc}$ be a binary $(n \times N)$ matrix that extracts the patch centered at position $(r,c)$.
%
With this notation, we formulate the (global) sparsity promoting function as
\begin{align}\label{eq:sumsum}
\sum \limits_{r \in \mathbf{r}} \sum \limits_{c \in \mathbf{c}} \sum \limits_{i = 1}^k ((\OP^\star \mathcal{P}_{rc}\mathbf{s} )_i^2+\nu)^{\frac{p}{2}},
\end{align}
which measures the overall approximated $\ell_p$-pseudo-norm of the considered analyzed image patches.
%
%
We compactly rewrite Equation \eqref{eq:sumsum} as
\begin{align}\label{eq:spfunc}
g(\OPF \mathbf{s}) :=  \sum \limits_{i=1}^{K} \left((\OPF \mathbf{s})_i^2+\nu\right)^{\frac{p}{2}},
\end{align}
with $K=k|\mathbf{r}||\mathbf{c}|$ and
\begin{align}\label{eq:fatoperator}
\OPF := \left[ \begin{array}{c}
\OP^\star \mathcal{P}_{r_1c_1} \\
\OP^\star \mathcal{P}_{r_1c_2} \\
\vdots \\
\OP^\star \mathcal{P}_{r_{|\mathbf{r}|}c_{|\mathbf{c}|}}
\end{array} \right] \in \R{K \times N}
\end{align}
being the \emph{global} analysis operator that expands the patch based one to the entire image.
We treat image boundary effects by employing constant padding, i.e.\ replicating the values at the image boundaries $\lfloor\frac{\sqrt{n}}{2}\rfloor$ times, where $\lfloor\cdot\rfloor$ denotes rounding to the smaller integer. Certainly, for image processing applications $\OPF$ is too large for being applied in terms of matrix vector multiplication. Fortunately, applying $\OPF$ and its transposed can be implemented efficiently using sliding window techniques, and the matrix vector notation is solely used for legibility.

According to \cite{rec:hawe:2012}, we exploit the fact that the range of pixel intensities is limited by a lower bound $b_l$ and an upper bound $b_u$. We enforce this bounding constraint by minimizing the differentiable function $\boldsymbol b(\mathbf{s}) := \sum\limits_{i=1}^N b(s_i)$, where $b$ is a penalty term given as
\begin{align}\label{eq:bound}
b(s) = \left\{ \begin{array}{cl}
|s - b_{u}|^2 &\mbox{ if $s \geq b_{u}$} \\
|s - b_{l}|^2 &\mbox{ if $s\leq b_{l}$} \\
0 &\mbox{ otherwise}
\end{array} \right..
\end{align}

Finally, combining the two constraints \eqref{eq:spfunc} and \eqref{eq:bound} with the data fidelity term, the analysis based image reconstruction problem is to solve
\begin{align}\label{eq:recovery}
\mathbf{s}^\star=\operatorname*{arg~min}_{\mathbf{s} \in \R{N}} \textstyle\frac{1}{2}\|\mathcal{A}\mathbf{s}-\mathbf{y}\|_2^2 + \boldsymbol b(\mathbf{s}) + \lambda g(\OPF \mathbf{s}).
\end{align}
Therein, $\lambda \in \R{+}$ balances between the sparsity of the solution's analysis coefficients and the solution's fidelity to the measurements. The measurement matrix $\mathcal{A} \in \R{m \times N}$ and the measurements $\mathbf{y} \in \R{m}$  are application dependent.

\section{Evaluation and Experiments}\label{sec:experments}
The first part of this section aims at answering the question of what is a good analysis operator for solving image reconstruction problems and relates the quality of an analysis operator with its mutual coherence and its condition number. This, in turn allows to select the optimal weighting parameters $\kappa$ and $\mu$ for GOAL. Using this parameters, we learn one general analysis operator $\OP^\star$ by GOAL, and compare its image denoising performance with other analysis approaches. In the second part, we employ this $\OP^\star$ unaltered for solving two classical image reconstruction tasks of image inpainting and single image super-resolution, and compare our results with the currently best analysis approach FoE \cite{mrf:roth:2009}, and state-of-the-art methods specifically designed for each respective application. 

\subsection{Global Parameters and Image Reconstruction}\label{subsec:global_para}
To quantify the reconstruction quality, as usual, we use the peak signal-to-noise ratio \linebreak $\PSNR = 10 \log( 255^2N/\sum_{i=1}^{N} (s_i-s_i^\star)^2 )$.
Moreover, we measure the quality using the Mean Structural SIMilarity Index ($\SSIM$) \cite{qu:wang:2004}, with the same set of parameters as originally suggested in \cite{qu:wang:2004}. Compared to $\PSNR$, the $\SSIM$ better reflects a human observer's visual impression of quality. It ranges between zero and one, with one meaning perfect image reconstruction.

Throughout all experiments, we fixed the size of the image patches to $(8\times 8)$, i.e. $n=64$. This is in accordance to the patch-sizes mostly used in the literature, and yields a good trade-off between reconstruction quality and numerical burden.
Images are reconstructed by solving the minimization problem \eqref{eq:recovery} via the conjugate gradient method proposed in \cite{rec:hawe:2012}. Considering the pixel intensity bounds, we used $b_l = 0$ and $b_u = 255$, which is the common intensity range in $8$-bit grayscale image formats. 
The sparsity promoting function \eqref{eq:spfunc} with $p=0.4$ and $\nu = 10^{-6}$ is used for both learning the analysis operator by GOAL, and reconstructing the images. Our patch based reconstruction algorithm as explained in Section \ref{sec:image_rec} achieves the best results for the maximum possible overlap $d_h=d_v=1$. The Lagrange multiplier $\lambda$ and the measurements matrix $\mathcal{A}$ depend on the application, and are briefly discussed in the respective subsections. 

\subsection{Analysis Operator Evaluation and Parameter Selection}
For evaluating the quality of an analysis operator and for selecting appropriate parameters for GOAL, we choose image denoising as the baseline experiment. The images to be reconstructed have artificially been corrupted by additive white Gaussian noise (AWGN) of varying standard deviation $\sigma_\textit{noise}$. This baseline experiment is further used to compare GOAL with other analysis operator learning methods.  We like to emphasize that the choice of image denoising as a baseline experiment is not crucial neither for selecting the learning parameters, nor for ranking the learning approaches. In fact, any other reconstruction task as discussed below leads to the same parameters and the same ranking of the different learning algorithms. 

\begin{figure}[htb]
\centering
\begin{minipage}[b]{\figwidth}
  \subfloat{\includegraphics[width=0.19\columnwidth,height=0.19\columnwidth]{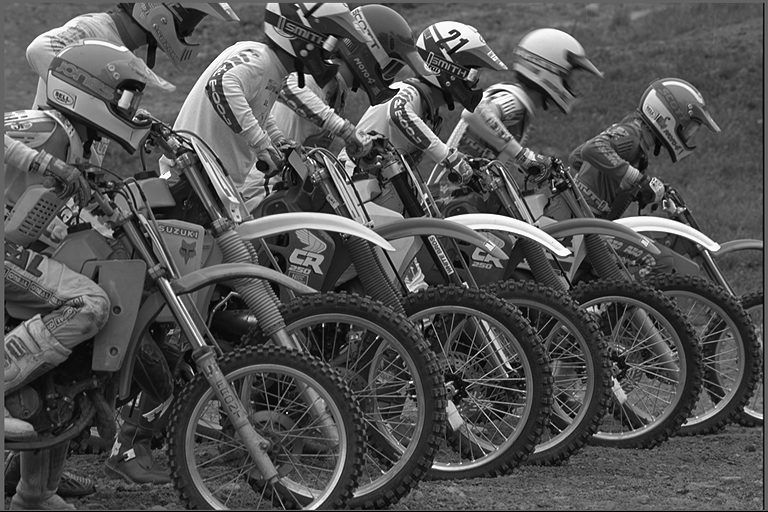}\label{fig:ti1}}
  \hfill
  \subfloat{\includegraphics[width=0.19\columnwidth,height=0.19\columnwidth]{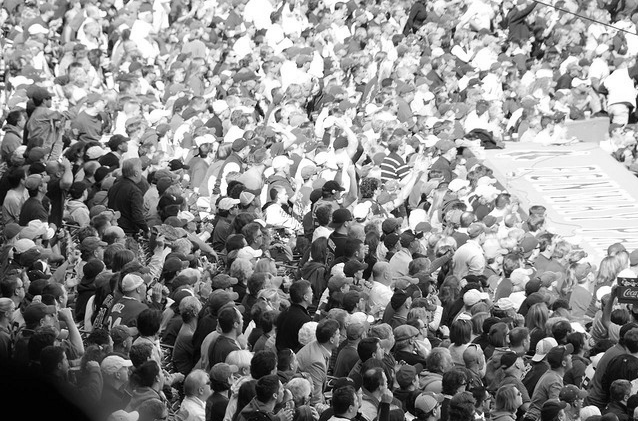}\label{fig:ti2}}
  \hfill
  \subfloat{\includegraphics[width=0.19\columnwidth,height=0.19\columnwidth]{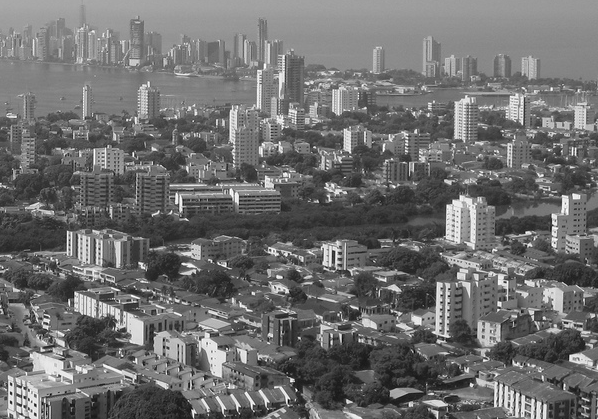}\label{fig:ti3}}
  \hfill
  \subfloat{\includegraphics[width=0.19\columnwidth,height=0.19\columnwidth]{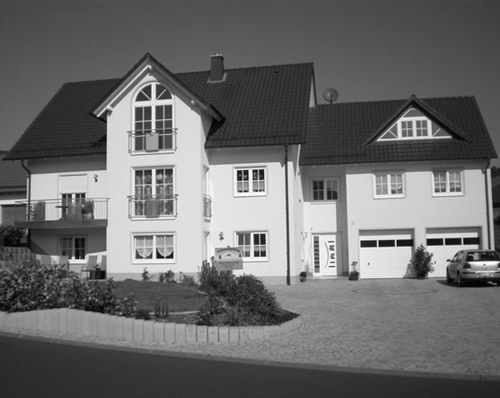}\label{fig:ti4}}
  \hfill
  \subfloat{\includegraphics[width=0.19\columnwidth,height=0.19\columnwidth]{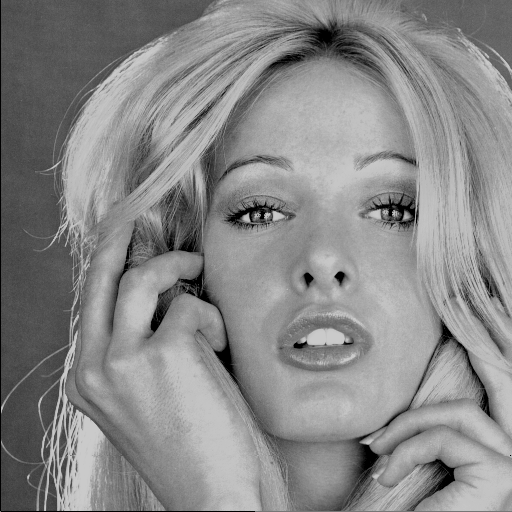}\label{fig:ti5}}
\end{minipage}
\caption{Five training images used for learning the analysis operator.}
\label{fig:trainingset}
\end{figure}

For image denoising, the measurement matrix $\mathcal{A}$ in Equation \eqref{eq:recovery} is the identity matrix. 
As it is common in the denoising literature, we assume the noise level $\sigma_\textit{noise}$ to be known and adjust $\lambda$ accordingly. From our experiments, we found that $\lambda = \frac{\sigma_\textit{noise}}{16}$ is a good choice. We terminate our algorithm after $6-30$ iterations depending on the noise level, i.e.\ the higher the noise level is the more iterations are required.
To find an optimal analysis operator, we learned several operators with varying values for $\mu, \kappa$, and $k$ and fixed all other parameters according to Subsection \ref{subsec:global_para}. Then, we evaluated their performance for the baseline task, which consists of denoising the five test images, each corrupted with the five noise levels as given in Table \ref{tb:psnrdenoise}. As the final performance measure we use the average $\PSNR$ of the 25 achieved results. 
The training set consisted of $M=200\;000$ image patches, each normalized to unit Euclidean norm, that have randomly been extracted from the five training images shown in Figure \ref{fig:trainingset}. Certainly, these images are not considered within any of the performance evaluations. Each time, we initialized GOAL with a random matrix having normalized rows. Tests with other initializations like an overcomplete DCT did not influence the final operator.

\begin{figure}[htb]
\centering
\begin{minipage}[b]{\columnwidth}
  \subfloat[][]{\includegraphics[width=0.48\columnwidth]{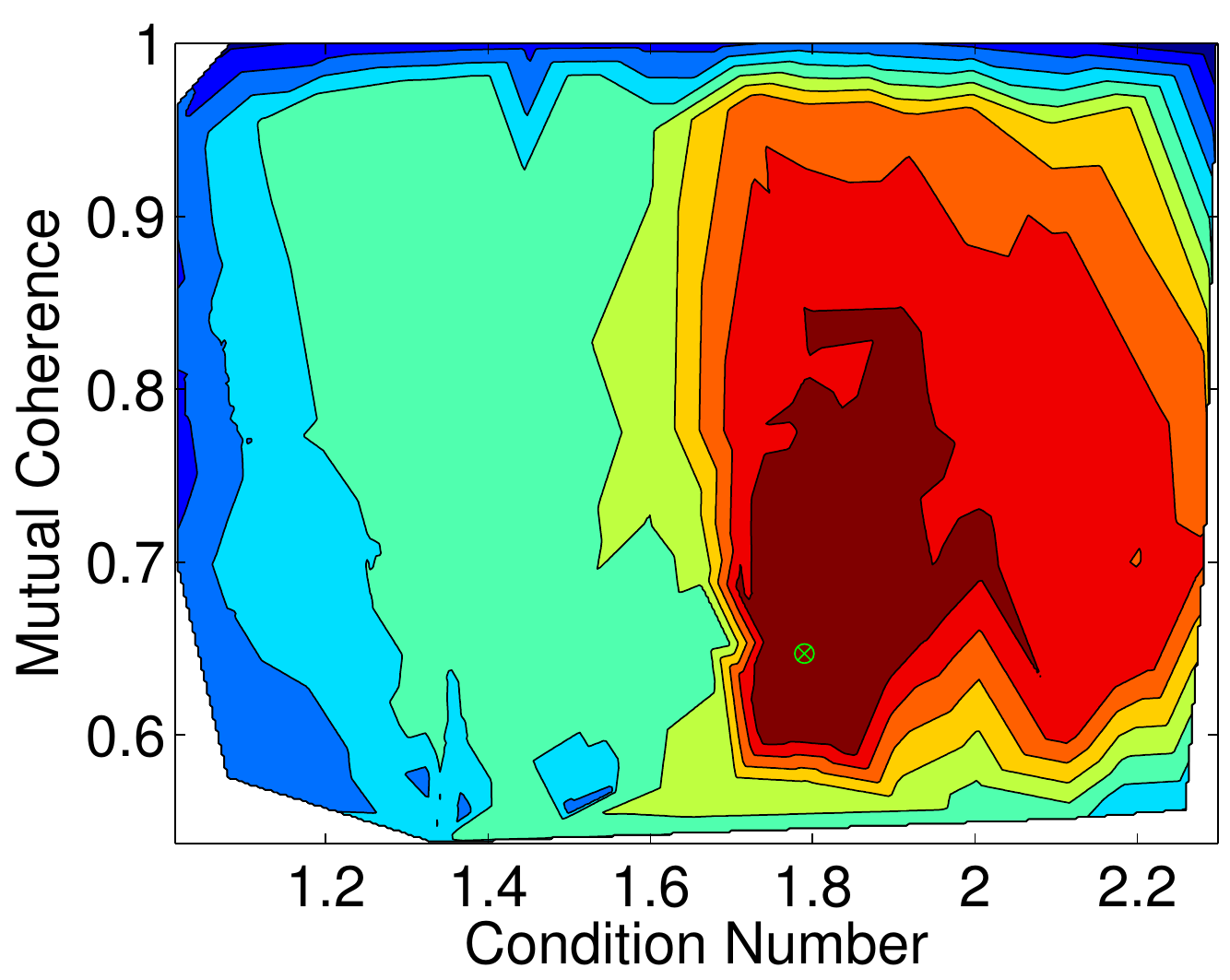}\label{fig:heatmap}}
  \hfill
  \subfloat[][]{\includegraphics[width=0.48\columnwidth]{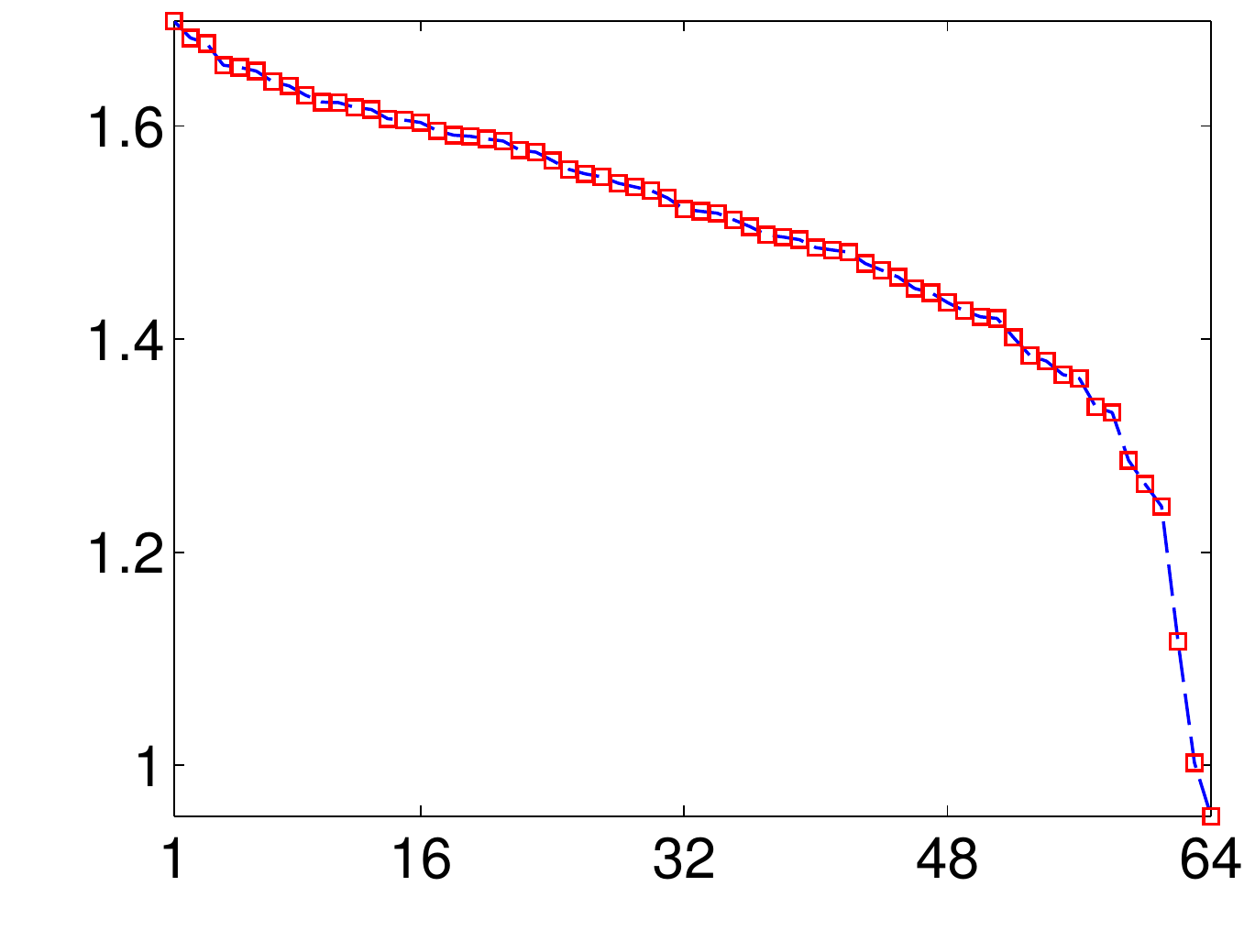}\label{fig:svds}}
 \end{minipage}
\caption{(a) Performance of $400$ analysis operators learned by GOAL in relation to their mutual coherence and their condition number. Color ranges from dark blue (worst) to dark red (best).  The green dot corresponds to the best performing operator $\OP^\star$.  (b) Singular values of $\OP^\star$}
\label{fig:heatsvd}
\end{figure}

\begin{figure}
\centering
\includegraphics[width=\figwidth]{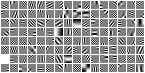}
\caption{Learned atoms of the analysis operator  $\OP^\star$. Each of the $128$ atoms is represented as a $8 \times 8$ square, where black corresponds to the smallest negative entry, gray is a zero entry, and white corresponds to the largest positive entry.}
\label{fig:aop}
\end{figure}

The results showed, that our approach clearly benefits from over-completeness. The larger we choose $k$, the better the operator performs with saturation starting at $k=2n$. Therefore, we fixed the number of atoms for all further experiments to $k=2n$.  Regarding $\kappa$ and $\mu$, note that by Lemma \ref{lemma:cond} and \ref{lemma:mc} these parameters influence the condition number and the mutual coherence of the learned operator.
Towards answering the question of what is a good condition number and mutual coherence for an analysis operator, Figure \ref{fig:heatsvd}\subref{fig:heatmap} shows the relative denoising performance of 400 operators learned by GOAL in relation to their mutual coherence and condition. We like to mention that according to our experiments, this relation is mostly independent from the degree of over-completeness. It is also interesting to notice that the best learned analysis operator is not a uniformly tight frame. The concrete values, which led to the best performing analysis operator $\OP^\star  \in \R{128 \times 64}$ in our experiments are $\kappa=9000$ and $\mu=0.01$. Its singular values are shown in Figure \ref{fig:heatsvd}\subref{fig:svds} and its atoms are visualized in Figure \ref{fig:aop}. This operator $\OP^\star$ remains unaltered throughout all following experiments in Subsections \ref{subsec:compare} -- \ref{subsec:IP}.  

\subsection{Comparison with Related Approaches} \label{subsec:compare}
The purpose of this subsection is to rank our approach among other analysis operator learning methods, and to compare its performance with state-of-the-art denoising algorithms. Concretely, we compare the denoising performance using $\OP^\star$ learned by GOAL with total-variation (TV) \cite{tv:dahl:2010} which is the currently best known analysis operator, with the recently proposed method AOL \cite{aol:yaghoobi:2011}, and with the currently best performing analysis operator FoE \cite{mrf:roth:2009}. Note that we used the same training set and dimensions for learning the operator by AOL as for GOAL. For FoE we used the same setup as originally suggested by the authors. 
Concerning the required computation time for learning an analysis operator, for this setting GOAL needs about $10$-minutes on an Intel Core i7 3.2 GHz quad-core with 8GB RAM. In contrast, AOL is approximately ten times slower, and FoE is the computationally most expensive method requiring several hours. All three methods are implemented in unoptimized Matlab code.
  
\begin{figure}[htb] %
\centering
\begin{minipage}[b]{\linewidth}
  \subfloat[][\centering GOAL, $\PSNR \ 30.44$dB $\SSIM \ 0.831$]{\includegraphics[width=0.49\textwidth,height=0.49\textwidth]{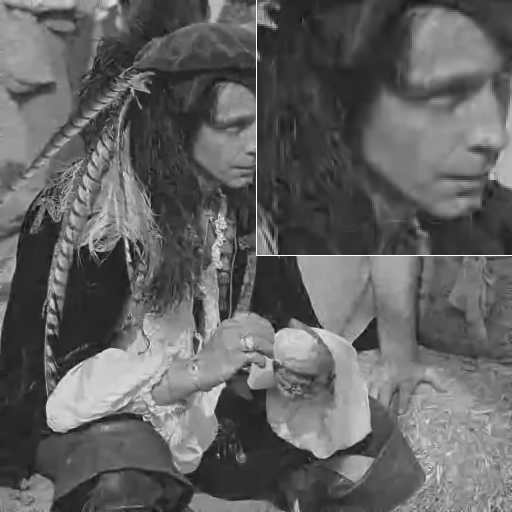}\label{fig:man}}
  \hfill
  \subfloat[][\centering AOL \cite{aol:yaghoobi:2011}, $\PSNR \ 27.33$dB $\SSIM \ 0.720$]{\includegraphics[width=0.49\textwidth,height=0.49\textwidth]{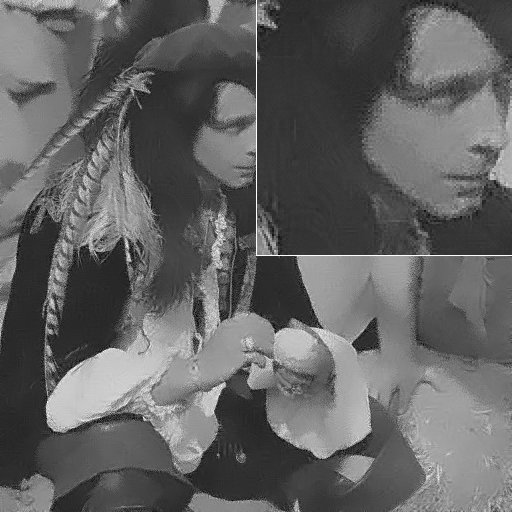}\label{fig:manbm3d}}
  \hfill
  \subfloat[][\centering TV \cite{tv:dahl:2010}, $\PSNR \ 29.63$dB $\SSIM \ 0.795$]{\includegraphics[width=0.49\textwidth,height=0.49\textwidth]{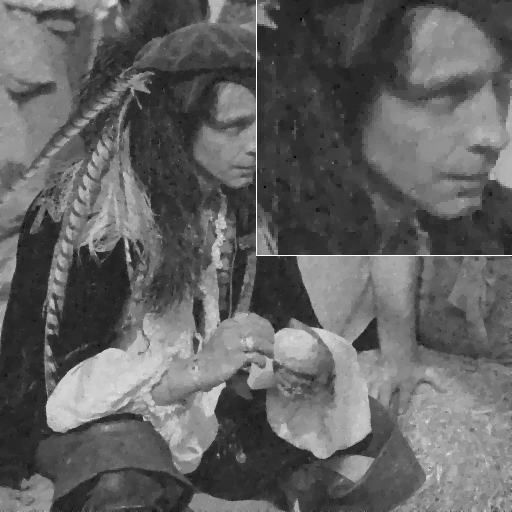}\label{fig:manksvd}}
  \hfill
  \subfloat[][\centering FoE \cite{mrf:roth:2009}, $\PSNR \ 29.75$dB $\SSIM \ 0.801$]{\includegraphics[width=0.49\textwidth,height=0.49\textwidth]{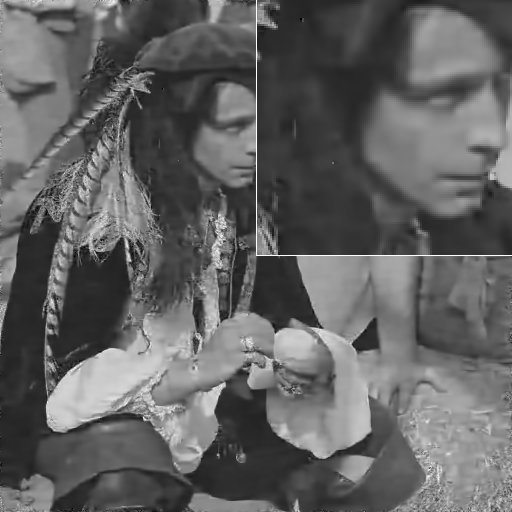}\label{fig:mantv}}
\end{minipage}
\caption{Images exemplarily showing the typical artifacts created by the four compared analysis operators for image denoising ("man" image, $\sigma_\textit{noise}=20$). For a better visualization a close up is provided for each image.}
\label{fig:denoiseexample}
\end{figure}

 The achieved results for the five test images and the five noise levels are given in Table \ref{tb:psnrdenoise}.
Our approach achieves the best results among the analysis methods both regarding $\PSNR$, and $\SSIM$. For a visual assessment, Figure \ref{fig:denoiseexample} exemplarily shows some denoising results achieved by the four analysis operators. 

To judge the analysis methods' denoising performance globally, we additionally give the results achieved by current state-of-the-art methods BM3D \cite{noise:dabov:2007} and K-SVD Denoising \cite{dl:elad:2006}, which are specifically designed for the purpose of image denoising.
In most of the cases our method performs slightly better than the K-SVD approach, especially for higher noise levels, and besides of the "barabara" image it is at most $\approx 0.5$dB worse than BM3D. This effect is due to the very special structure of the "barbara" image that rarely occurs in natural images, which are smoothed by the learned operator.

\begin{table*}
\centering
\caption{Achieved $\textit{PSNR}$ in decibels (dB) and $\SSIM$ for denoising five test images corrupted by five noise levels. Each cell contains the achieved results for the respective image with six different algorithms, which are: Top left GOAL, top right AOL \cite{aol:yaghoobi:2012}, middle left TV \cite{tv:dahl:2010}, middle right FoE \cite{mrf:roth:2009}, bottom left K-SVD denoising \cite{dl:elad:2006} bottom right BM3D \cite{noise:dabov:2007}}
\resizebox{\textwidth}{!}{
\begin{tabular}{||r||cc"cc||cc"cc||cc"cc||cc"cc||cc"cc||}
\hline  \hline
& \multicolumn{4}{c||}{\textbf{lena}} & \multicolumn{4}{c||}{\textbf{barbara}} & \multicolumn{4}{c||}{\textbf{man}} & \multicolumn{4}{c||}{\textbf{boat}} & \multicolumn{4}{c||}{\textbf{couple}} \\ \cline{2-21}
 $\sigma_\textit{noise} \ / \ \textit{PSNR}$  & \multicolumn{2}{c"}{$\PSNR$} & \multicolumn{2}{c||}{$\SSIM$} & \multicolumn{2}{c"}{$\PSNR$} & \multicolumn{2}{c||}{$\SSIM$} & \multicolumn{2}{c"}{$\PSNR$} & \multicolumn{2}{c||}{$\SSIM$} & \multicolumn{2}{c"}{$\PSNR$} & \multicolumn{2}{c||}{$\SSIM$} & \multicolumn{2}{c"}{$\PSNR$} & \multicolumn{2}{c||}{$\SSIM$}\\
\hline  \hline
$5 \ / \ 34.15$ & 38.65 & 36.51 & 0.945 & 0.924 & 37.96 & 35.95 & 0.962 & 0.944 & 37.77 & 35.91 & 0.954 & 0.932 & 37.09 & 35.77 & 0.938 & 0.926 & 37.43 & 35.55 & 0.951 & 0.932\\
 & 37.65 & 38.19 & 0.936 & 0.938 & 35.56 & 37.25 & 0.948 & 0.958 & 36.79 & 37.45 & 0.944 & 0.949 & 36.17 & 36.33 & 0.925 & 0.917 & 36.26 & 37.06 & 0.940 & 0.944\\
 & 38.48 & 38.45 & 0.944 & 0.942 & 38.12 & 38.27 & 0.964 & 0.964 & 37.51 & 37.79 & 0.952 & 0.954 & 37.14 & 37.25 & 0.939 & 0.938 & 37.24 & 37.14 & 0.950 & 0.951\\
\hline  \hline
$10 \ / \ 28.13$ & 35.58 & 32.20 & 0.910 & 0.856 & 33.98 & 31.27 & 0.930 & 0.883 & 33.88 & 31.33 & 0.907 & 0.851 & 33.72 & 31.24 & 0.883 & 0.842 & 33.75 & 30.87 & 0.903 & 0.844\\
 & 34.24 & 35.12 & 0.890 & 0.901 & 30.84 & 32.91 & 0.886 & 0.923 & 32.90 & 33.44 & 0.884 & 0.893 & 32.54 & 33.23 & 0.863 & 0.868 & 32.32 & 33.37 & 0.878 & 0.889\\
 & 35.52 & 35.79 & 0.910 & 0.915 & 34.56 & 34.96 & 0.936 & 0.942 & 33.64 & 33.97 & 0.901 & 0.907 & 33.68 & 33.91 & 0.883 & 0.887 & 33.62 & 33.86 & 0.901 & 0.909\\
\hline  \hline
$20 \ / \ 22.11$ & 32.63 & 28.50 & 0.869 & 0.772 & 30.17 & 27.26 & 0.880 & 0.791 & 30.44 & 27.33 & 0.831 & 0.720 & 30.62 & 27.16 & 0.819 & 0.711 & 30.39 & 26.95 & 0.833 & 0.727\\
 & 31.09 & 31.97 & 0.827 & 0.856 & 26.79 & 28.39 & 0.773 & 0.849 & 29.63 & 29.75 & 0.795 & 0.801 & 29.30 & 29.96 & 0.778 & 0.793 & 28.87 & 29.77 & 0.783 & 0.807\\
 & 32.39 & 32.98 & 0.861 & 0.875 & 30.87 & 31.78 & 0.881 & 0.905 & 30.17 & 30.59 & 0.814 & 0.833 & 30.44 & 30.89 & 0.805 & 0.825 & 30.08 & 30.68 & 0.817 & 0.847\\
\hline  \hline
$25 \ / \ 20.17$ & 31.65 & 27.47 & 0.854 & 0.742 & 29.05 & 26.08 & 0.856 & 0.750 & 29.43 & 26.28 & 0.801 & 0.677 & 29.61 & 26.08 & 0.792 & 0.671 & 29.32 & 25.81 & 0.802 & 0.679\\
 & 30.05 & 30.87 & 0.796 & 0.836 & 25.73 & 27.05 & 0.724 & 0.813 & 28.66 & 28.62 & 0.759 & 0.761 & 28.32 & 28.87 & 0.744 & 0.758 & 27.87 & 28.57 & 0.746 & 0.767\\
 & 31.33 & 32.02 & 0.842 & 0.859 & 29.59 & 30.72 & 0.850 & 0.887 & 29.14 & 29.62 & 0.780 & 0.804 & 29.36 & 29.92 & 0.772 & 0.801 & 28.92 & 29.65 & 0.780 & 0.820\\
\hline  \hline
$30 \ / \ 18.59$ & 30.86 & 26.50 & 0.839 & 0.717 & 27.93 & 24.95 & 0.818 & 0.706 & 28.64 & 25.30 & 0.774 & 0.638 & 28.80 & 25.07 & 0.769 & 0.630 & 28.46 & 24.79 & 0.780 & 0.633\\
 & 29.40 & 30.00 & 0.786 & 0.823 & 24.91 & 25.97 & 0.690 & 0.787 & 27.95 & 27.85 & 0.736 & 0.740 & 27.56 & 28.01 & 0.720 & 0.737 & 27.09 & 27.70 & 0.715 & 0.743\\
 & 30.44 & 31.22 & 0.823 & 0.843 & 28.56 & 29.82 & 0.821 & 0.868 & 28.30 & 28.87 & 0.750 & 0.780 & 28.48 & 29.13 & 0.744 & 0.779 & 27.95 & 28.81 & 0.746 & 0.795\\
\hline  \hline
\end{tabular}
}
\label{tb:psnrdenoise}
\end{table*}

\subsection{Image Inpainting}
In image inpainting as originally proposed in \cite{ip:bertalmio:2000}, the goal is to fill up a set of damaged or disturbing pixels such that the resulting image is visually appealing. This is necessary for the restoration of damaged photographs, for removing disturbances caused by e.g. defective hardware, or for deleting unwanted objects. Typically, the positions of the pixels to be filled up are given a priori. In our formulation, when $N-m$ pixels must be inpainted, this leads to a binary $m \times N$ dimensional measurements matrix $\mathcal{A}$, where each row contains exactly one entry equal to one. Its position corresponds to a pixel with known intensity. Hence, $\mathcal{A}$ reflects the available image information. Regarding $\lambda$, it can be used in a way that our method simultaneously inpaints missing pixels and denoises the remaining ones.

\begin{figure}[htb]
\centering
\begin{minipage}[b]{\linewidth}
  \subfloat[][\centering Masked $90\%$ missing pixels.]{\includegraphics[width=0.49\linewidth,height=0.49\linewidth]{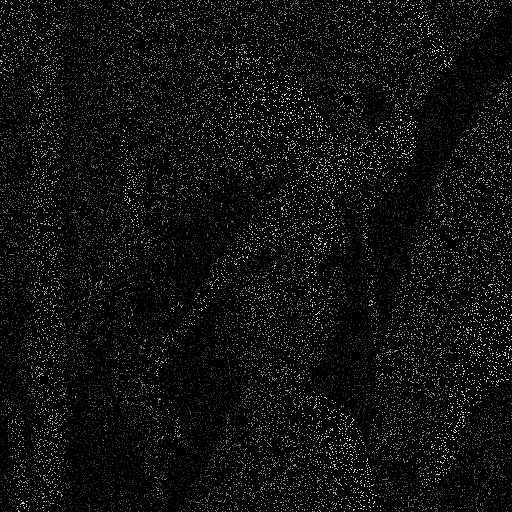}\label{fig:plena10}}
  \hfill
  \subfloat[][\centering Inpainted image GOAL, $\textit{PSNR} \ 28.57$dB $\SSIM \ 0.840$.]{\includegraphics[width=0.49\linewidth,height=0.49\linewidth]{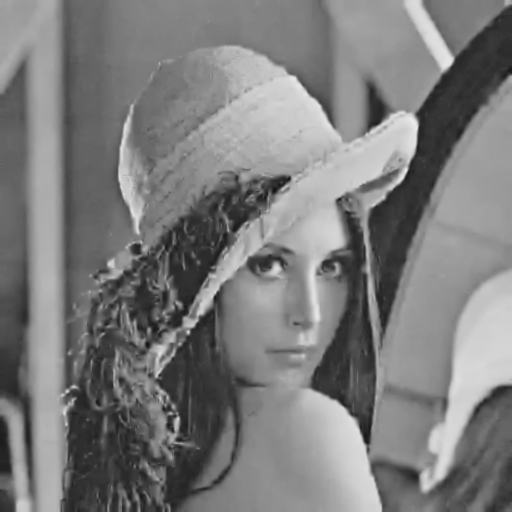}\label{fig:rlenaGOAL}}
  \hfill
  \subfloat[][\centering Inpainted image FoE, $\textit{PSNR} \ 28.06$dB $\SSIM  \ 0.822$.]{\includegraphics[width=0.49\linewidth,height=0.49\linewidth]{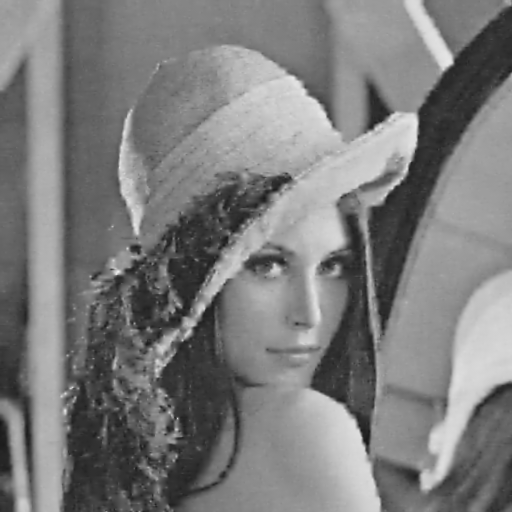}\label{fig:rlenaFOE}}
  \hfill
  \subfloat[][\centering Inpainted image \cite{ip:zhou:2012}, $\textit{PSNR} \ 27.63$dB $\SSIM  \ 0.804$.]{\includegraphics[width=0.49\linewidth,height=0.49\linewidth]{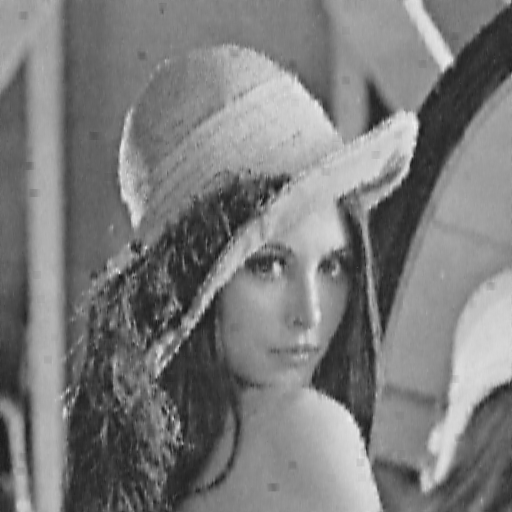}\label{fig:rlenaGT}}
\end{minipage}
\caption{Results for reconstructing the "lena" image from $10\%$ of all pixels using $\OP^\star$ learny by GOAL, FoE, and \cite{ip:zhou:2012}.}
\label{fig:syninpaint}
\end{figure}

As an example for image inpainting, we disturbed some ground-truth images artificially by removing $N-m$ pixels randomly distributed over the entire image as exemplary shown in Figure \ref{fig:syninpaint}\subref{fig:plena10}. In that way, the reconstruction quality can be judged both visually and quantitatively. We assumed the data to  be free of noise, and empirically selected $\lambda = 10^{-2}$. In Figure \ref{fig:syninpaint}, we show exemplary results for reconstructing the "lena" image from $10\%$ of all pixels using GOAL, FoE, and the recently proposed synthesis based method \cite{ip:zhou:2012}. Table \ref{tb:psnrip} gives a comparison of further images and further number of missing pixels. It can be seen that our methods performs best independent of the configuration.

\begin{table}
\centering
\caption{Results achieved for inpainting three test images with varying number of missing pixels using three different methods. In each cell, the $\PSNR$ in dB and the $\SSIM$ are given for GOAL (top), FoE \cite{mrf:roth:2009}(middle), and method \cite{ip:zhou:2012} (bottom).}
\resizebox{\figwidth}{!}{
\begin{tabular}{||r||c|c||c|c||c|c||}
\hline  \hline
$\%$ of missing pixels & \multicolumn{2}{c||}{\textbf{lena}} & \multicolumn{2}{c||}{\textbf{boat}} & \multicolumn{2}{c||}{\textbf{man}} \\ \cline{2-7}
 & $\PSNR$ & $\SSIM$ & $\PSNR$ & $\SSIM$ & $\PSNR$ & $\SSIM$\\
\hline  \hline
$0.90\%$ & 28.57 & 0.840  & 25.61 & 0.743  & 26.35 & 0.755 \\ \cline{2-7}
 & 28.06 & 0.822  & 25.14 & 0.719  & 26.23 & 0.747 \\ \cline{2-7}
 & 27.63 & 0.804  & 24.80 & 0.683  & 25.56 & 0.715 \\ 
 \hline \hline 
$0.80\%$ & 31.82 & 0.895  & 28.55 & 0.833  & 28.93 & 0.847 \\ \cline{2-7}
 & 31.09 & 0.880  & 27.76 & 0.804  & 28.51 & 0.836 \\ \cline{2-7}
 & 30.95 & 0.878  & 27.80 & 0.804  & 28.24 & 0.821 \\ 
 \hline \hline 
$0.50\%$ & 37.75 & 0.956  & 34.47 & 0.936  & 34.12 & 0.947 \\ \cline{2-7}
 & 36.70 & 0.947  & 33.17 & 0.907  & 33.49 & 0.940 \\ \cline{2-7}
 & 36.75 & 0.943  & 33.77 & 0.918  & 33.27 & 0.934 \\ 
 \hline \hline 
$0.20\%$ & 43.53 & 0.985  & 41.04 & 0.982  & 40.15 & 0.985 \\ \cline{2-7}
 & 42.29 & 0.981  & 38.45 & 0.963  & 39.15 & 0.982 \\ \cline{2-7}
 & 40.77 & 0.965  & 39.45 & 0.966  & 39.06 & 0.977 \\ 
 \hline \hline 
\end{tabular}
}
\label{tb:psnrip}
\end{table}

\subsection{Single Image Super-Resolution}\label{subsec:IP}
In single image super-resolution (SR), the goal is to reconstruct a high resolution image $\mathbf{s} \in \R{N}$ from an observed low resolution image $\mathbf{y} \in \R{m}$. In that, $\mathbf{y}$ is assumed to be a blurred and downsampled version of  $\mathbf{s}$. Mathematically, this process can be formulated as $\mathbf{y}=\mathcal{D}\mathcal{B}\mathbf{s} + \mathbf{e}$ where $\mathcal{D} \in \R{m \times N}$ is a decimation operator and $\mathcal{B} \in \R{N \times N}$ is a blur operator. Hence, the measurement matrix is given by $\mathcal{A}=\mathcal{D}\mathcal{B}$. In the ideal case, the exact blur kernel is known or an estimate is given. Here, we consider the more realistic case of an unknown blur kernel. Therefore, to apply our approach for magnifying an image by a factor of $d$ in both vertical and horizontal dimension, we model the blur via a Gaussian kernel of dimension $(2d - 1) \times (2d - 1)$ and with standard deviation $\sigma_\textit{blur} = \frac{d}{3}$.

For our experiments, we artificially created a low resolution image by downsampling a ground-truth image by a factor of $d$ using bicubic interpolation. Then, we employed bicubic interpolation, FoE, the method from \cite{sr:yang:2010}, and GOAL to magnify this low resolution image by the same factor $d$. This upsampled version is then compared with the original image in terms of $\PSNR$ and $\SSIM$. In Table \ref{tb:results_sr}, we present the results for upsampling the respective images by $d=3$. The presented results show that our method outperforms the current state-of-the-art. We want to emphasize that the blur kernel used for downsampling is different from the blur kernel used in our upsampling procedure.

\begin{table*}
\centering
\caption{The results in terms of $\PSNR$ and $\SSIM$ for upsampling the seven test images by a factor of $d=3$ using five different algorithms GOAL, FoE \cite{mrf:roth:2009}, method \cite{sr:yang:2010}, and Bicubic interpolation.}
\resizebox{\textwidth}{!}{
\begin{tabular}{||r||c|c||c|c||c|c||c|c||c|c||c|c||c|c||}
\hline  \hline
Method & \multicolumn{2}{c||}{\textbf{face}} & \multicolumn{2}{c||}{\textbf{august}} & \multicolumn{2}{c||}{\textbf{barbara}} & \multicolumn{2}{c||}{\textbf{lena}} & \multicolumn{2}{c||}{\textbf{man}} & \multicolumn{2}{c||}{\textbf{boat}} & \multicolumn{2}{c||}{\textbf{couple}} \\ \cline{2-15}
 & $\PSNR$ & $\SSIM$ & $\PSNR$ & $\SSIM$ & $\PSNR$ & $\SSIM$ & $\PSNR$ & $\SSIM$ & $\PSNR$ & $\SSIM$ & $\PSNR$ & $\SSIM$ & $\PSNR$ & $\SSIM$\\
\hline  \hline
GOAL  & 32.37 & 0.801  & 23.28 & 0.791  & 24.42 & 0.731  & 32.36 & 0.889  & 29.48 & 0.837  & 28.25 & 0.800  & 27.79 & 0.786 \\
\hline  \hline
FoE  & 32.19 & 0.797  & 22.95 & 0.782  & 24.30 & 0.727  & 31.82 & 0.885  & 29.17 & 0.832  & 28.00 & 0.797  & 27.64 & 0.782 \\
\hline  \hline
Method \cite{sr:yang:2010}  & 32.16 & 0.795  & 22.90 & 0.771  & 24.25 & 0.719  & 32.00 & 0.881  & 29.29 & 0.829  & 28.04 & 0.793  & 27.56 & 0.778 \\
\hline  \hline
Bicubic  & 31.57 & 0.771  & 22.07 & 0.724  & 24.13 & 0.703  & 30.81 & 0.863  & 28.39 & 0.796  & 27.18 & 0.759  & 26.92 & 0.743 \\
\hline  \hline
\end{tabular}
}
\label{tb:results_sr}
\end{table*}

\begin{figure}[htb]
\centering
\begin{minipage}[b]{\linewidth}
  \subfloat[][\centering Original image "august".]{\includegraphics[width=0.47\linewidth]{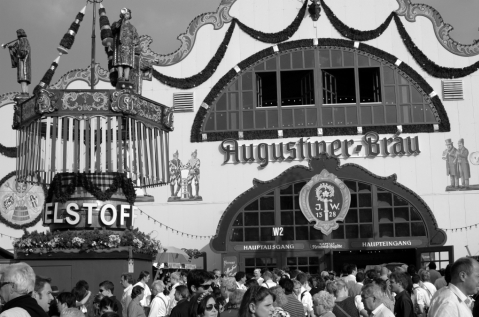}\label{fig:august}}
  \hfill
  \subfloat[][\centering Noisy low resolution image.]{\includegraphics[width=0.47\linewidth]{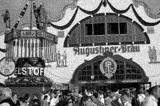}\label{fig:augustnoisy}}
  \hfill
  \subfloat[][\centering Bicubic Interpolation, $\PSNR \ 21.63$dB $\SSIM \ 0.653$.]{\includegraphics[width=0.47\linewidth]{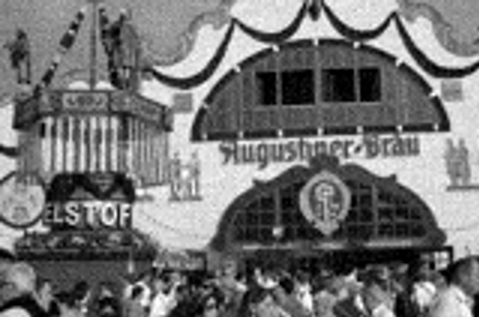}\label{fig:augustbi}}
  \hfill
  \subfloat[][\centering Method \cite{sr:yang:2010}, $\PSNR \ 22.07$dB $\SSIM \ 0.663$.]{\includegraphics[width=0.47\linewidth]{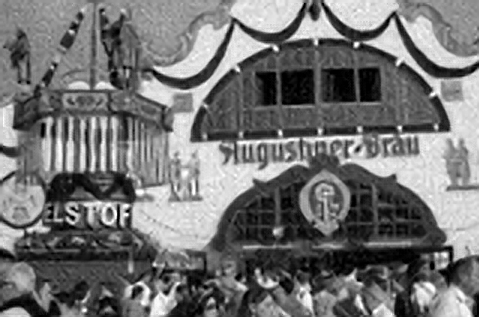}\label{fig:augustyi}}
  \hfill
  \subfloat[][\centering FoE \cite{mrf:roth:2009}, $\PSNR \ 22.17$dB $\SSIM \ 0.711$.]{\includegraphics[width=0.47\linewidth]{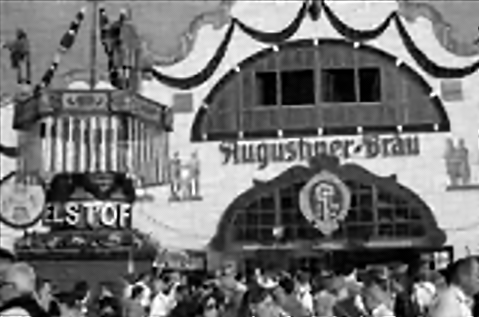}\label{fig:augusttv}}
  \hfill
  \subfloat[][\centering GOAL, $\PSNR \ 22.45$dB $\SSIM \ 0.726$.]{\includegraphics[width=0.47\linewidth]{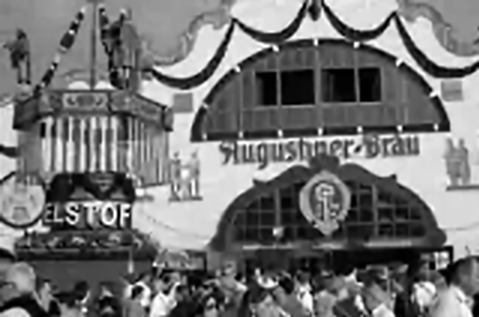}\label{fig:augustwe}}
  \hfill
\end{minipage}
\caption{Single image super-resolution results of four algorithms on noisy data, for magnifying a low resolution image by a factor of three together with the corresponding $\PSNR$ and $\SSIM$. The low resolution image has been corrupted by AWGN with $\sigma_\textit{noise}=8$.}
\label{fig:noisysri}
\end{figure}

Note that many single image super-resolution algorithms rely on clean noise free input data, whereas the general analysis approach as formulated in Equation \eqref{eq:recovery} naturally handles noisy data, and is able to perform simultaneous upsampling and denoising. In Figure \ref{fig:noisysri} we present the result for simultaneously denoising and upsampling a low resolution version of the image "august" by a factor of $d=3$, which has been corrupted by AWGN with $\sigma_\textit{noise}=8$. As it can be seen, our method produces the best results both visually and quantitatively, especially regarding the $\SSIM$. Due to high texture this image is hard to upscale even when no noise is present, see the second column of Table \ref{tb:results_sr}. Results obtained for other images confirm this good performance of GOAL but are not presented here due to space limitation.

\section{Conclusion}
This paper deals with the topic of learning an analysis operator from example image patches, and how to apply it for solving inverse problems in imaging. To learn the operator, we motivate an $\ell_p$-minimization on the set of full-rank matrices with normalized columns. A geometric conjugate gradient method on the oblique manifold is suggested to solve the arising optimization task. Furthermore, we give a partitioning invariant method for employing the local patch based analysis operator such that globally consistent reconstruction results are achieved. For the famous tasks of image denoising, image inpainting, and single image super-resolution, we provide promising results that are competitive with and even outperform current state-of-the-art techniques. Similar as for the synthesis signal reconstruction model with dictionaries, we expect that depending on the application at hand, the performance of the analysis approach can be further increased by learning the particular operator with regard to the specific problem, or employing a specialized training set.


%

%

\ifCLASSOPTIONcaptionsoff
  \newpage
\fi



\bibliographystyle{IEEEtran}
\bibliography{literatur}

%
%

\end{document}